%% file: main.tex
\documentclass[10pt,twocolumn,letterpaper]{article}

\usepackage[pagenumbers]{cvpr} 

\input{preamble}

\definecolor{cvprblue}{rgb}{0.21,0.49,0.74}
\usepackage[pagebackref,breaklinks,colorlinks,allcolors=cvprblue]{hyperref}
\usepackage{xcolor}  
\usepackage{float}
\usepackage{pifont}   
\usepackage{subcaption}
\usepackage[table]{xcolor}
\usepackage{svg}
\usepackage{tcolorbox}
\usepackage{enumitem}
\tcbuselibrary{listings,breakable,skins,xparse}

\def\paperID{1601} 
\def\confName{CVPR}
\def\confYear{2026}

\title{OmniScience: A Large-scale Multi-modal Dataset for Scientific Image Understanding}

\author{
Haoyi Tao\footnotemark[1] \quad
Chaozheng Huang\footnotemark[1] \quad
Nan Wang\footnotemark[1] \quad
Han Lyu\footnotemark[1] \\
Linfeng Zhang \quad
Guolin Ke \quad
Xi Fang\footnotemark[2] \\[2mm]
DP Technology\\
{\tt\small \{taohaoyi, huangchaozheng, wangnan01, lvhan, zhanglf, kegl, fangxi\}@dp.tech}
}

\begin{document}
\maketitle

\renewcommand{\thefootnote}{\fnsymbol{footnote}}
\footnotetext[1]{These authors contributed equally to this work.}
\footnotetext[2]{Corresponding author.}
\renewcommand{\thefootnote}{\arabic{footnote}}

\input{sec/0_abstract}    
\input{sec/1_introduction}
\input{sec/2_related_work}

\input{sec/3_data_curation}
\input{sec/4_experiment}
\input{sec/5_conclusion}

{
    \small
    \bibliographystyle{ieeenat_fullname}
    \bibliography{main}
}

\input{sec/appendix}

\end{document}

%% file: preamble.tex
\usepackage{pifont}
\usepackage{xcolor}

%% file: sec/0_abstract.tex
\begin{abstract}
Multimodal Large Language Models (MLLMs) demonstrate strong performance on natural image understanding, yet exhibit limited capability in interpreting scientific images, including but not limited to schematic diagrams, experimental characterizations, and analytical charts. This limitation is particularly pronounced in open-source MLLMs. The gap largely stems from existing datasets with limited domain coverage, coarse structural annotations, and weak semantic grounding. 
We introduce OmniScience, a large-scale, high-fidelity multi-modal dataset comprising 1.5 million figure–caption–context triplets, spanning more than 10 major scientific disciplines.
To obtain image caption data with higher information density and accuracy for multi-modal large-model training, we develop a dynamic model-routing re-captioning pipeline that leverages state-of-the-art multi-modal large language models (MLLMs) to generate dense, self-contained descriptions by jointly synthesizing visual features, original figure captions, and corresponding in-text references authored by human scientists. The pipeline is further reinforced with rigorous quality filtering and alignment with human expert judgments, ensuring both factual accuracy and semantic completeness, and boosts the image–text multi-modal similarity score from $0.769$ to $0.956$. We further propose a caption QA protocol as a proxy task for evaluating visual understanding. Under this setting, Qwen2.5-VL-3B model finetuned on OmniScience show substantial gains over baselines, achieving $+0.378$ on MM-MT-Bench and $+0.140$ on MMMU.
We expect OmniScience to be a cornerstone for future AI scientists, bridging the gap in large-scale scientific image comprehension. The data is available at \url{https://huggingface.co/datasets/UniParser/OmniScience}.
\end{abstract}

%% file: sec/1_introduction.tex
\section{Introduction}
\label{sec:intro}

\begin{figure*}[ht]
\centering
\includegraphics[width=0.9\textwidth]{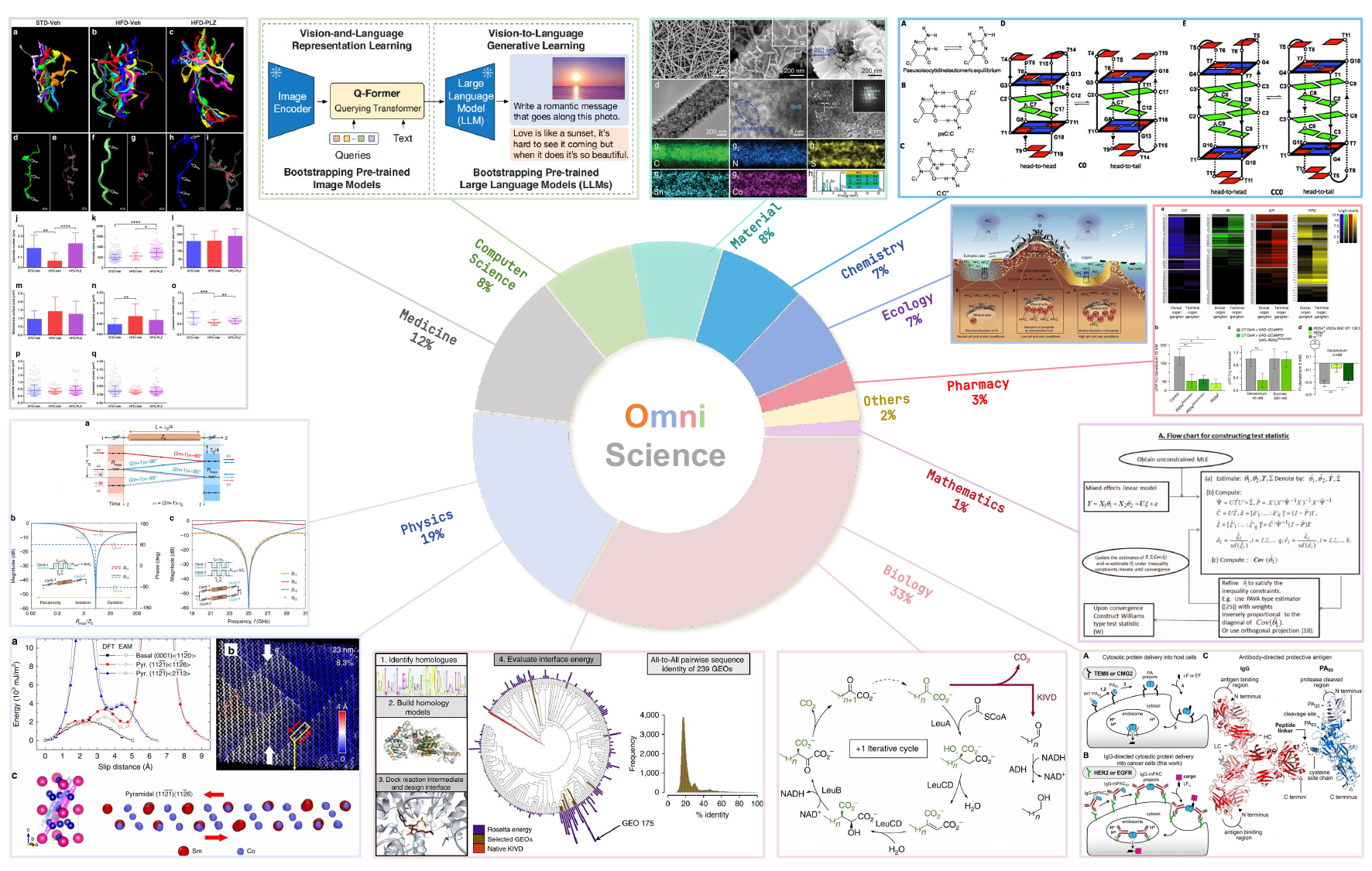} 
\caption{Disciplinary composition and visual diversity of OmniScience. The central pie chart summarizes the distribution of figures across major scientific fields, while surrounding examples highlight rich visual heterogeneity across domains. Together, they demonstrate the broad disciplinary coverage and diverse visual distributions of OmniScience.}
\label{fig:sample_image2}
\end{figure*}

\begin{figure*}[ht]
\centering
\includegraphics[width=0.9\textwidth]{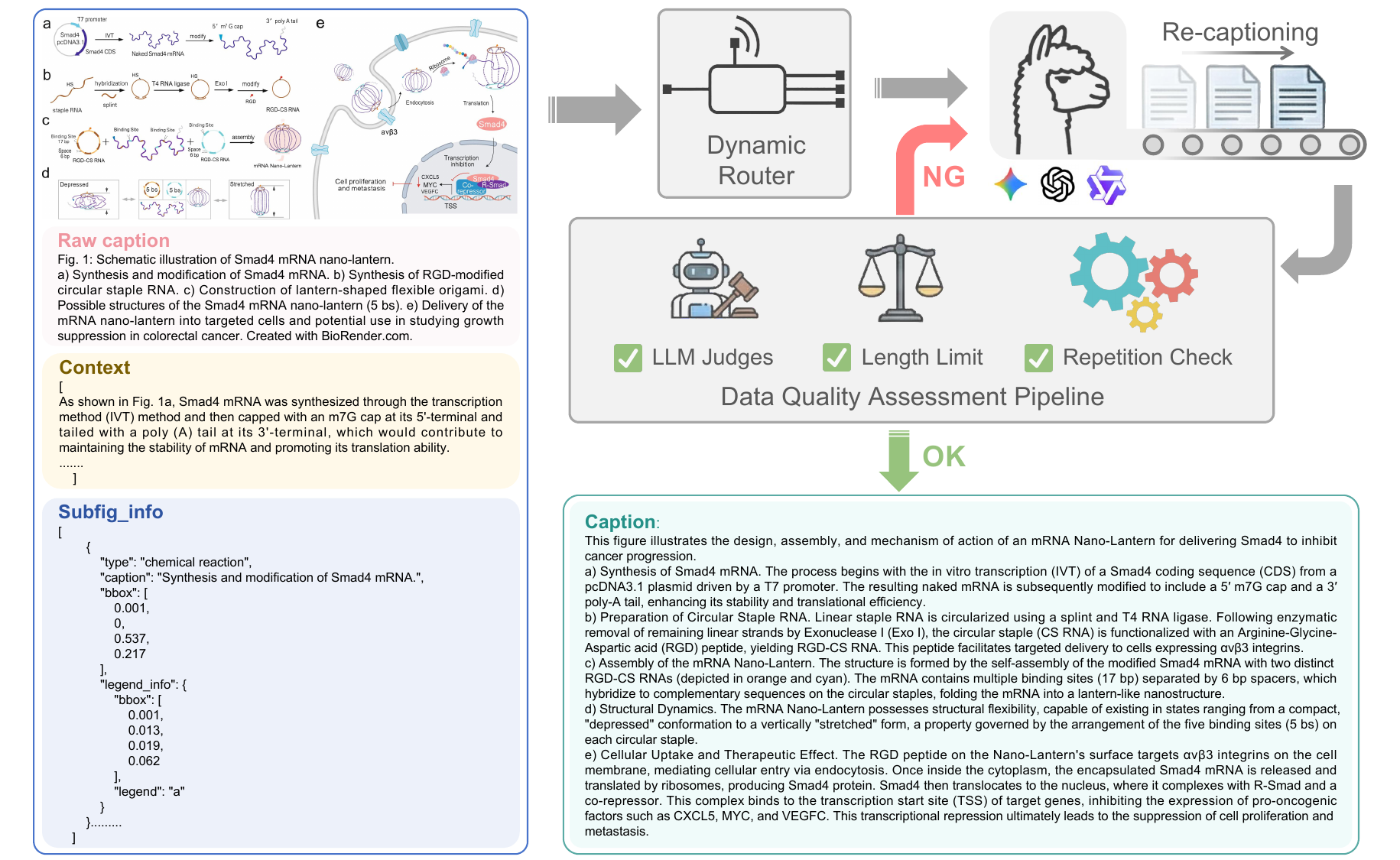}
\caption{Overview of OmniScience data construction. The example is taken from Hu \etal~\cite{hu2023lantern}. Each scientific figure is represented as an image–caption–context triplet, where the raw caption is paired with its corresponding context paragraph(s) extracted from the source article to provide surrounding textual grounding. Fine-grained subfigure annotations specify each panel’s bounding box, sub-caption, and semantic type (e.g., “chemical reaction”, “microscopy”, etc.). Additionally, a recaption field provides a refined summary that integrates visual and textual cues to fully convey the scientific meaning of the figure.}
\label{fig:sample_image}
\end{figure*}

The automated interpretation of scientific literature emerges as a grand challenge for artificial intelligence, offering the potential to fundamentally accelerate the pace of research and discovery~\cite{spangler2014automated,de2023artificial,krenn2022scientific,gottweis2025towards,sternlicht2025chimera,akujuobi2024link,liu2025literature}. A key frontier in this pursuit hinges upon the development of Multi-modal Large Language Models (MLLMs) capable of deciphering the dense, multi-modal information encapsulated within scientific figures. These visual representations, ranging from intricate schematic diagrams and microscopy images to complex analytical plots, serve as the primary medium for articulating experimental results and distilling core scientific insights. Accurate comprehension of such content often demands specialized knowledge and rigorous reasoning comparable to that of human experts at the doctoral level, posing a significant barrier for current general-purpose models.

Despite the rapid evolution of MLLMs, their efficacy in complex scientific domains remains constrained by critical shortcomings in existing datasets and benchmarks. This performance bottleneck is particularly evident in open-source models, which often lack exposure to specialized, high-fidelity scientific data. Early research in this area primarily addressed simplistic data visualizations, such as bar charts and scatter plots, frequently relying on synthetic or narrowly scoped datasets~\cite{dvqa,figureqa,plotqa}. While subsequent efforts expanded domain coverage, they often utilized pre-print repositories or included relatively straightforward figures that fail to mirror the structural complexity of high-end scientific publishing~\cite{chartqa,wang2024charxiv,hsu2021scicap,yang2024scicap+,mmmu, mpaper, chen2024we}. More recently, initiatives such as Multimodal Arxiv~\cite{arxivqa} and MMSCI~\cite{li2024mmsci} have significantly advanced the scale and diversity of scientific figure collections. MMSCI, in particular, enhances data fidelity by extracting figures from peer-reviewed sources via structured HTML. However, such structured data represent only a fraction of the scientific corpus; the vast majority of scientific knowledge remains ``locked'' within heterogeneous PDF layouts, where reliably extracting high-quality figure--text pairs at scale remains a formidable obstacle for the community.

To bridge these gaps, we introduce \textbf{OmniScience}, a large-scale, high-fidelity multimodal scientific dataset engineered through a paradigm of rigorous curation and systematic quality control. Our data acquisition pipeline specifically targets high-impact open-access publications (averaging an impact factor $>$ 12), as these sources represent a vast repository of expert-level scientific insights typically encapsulated in unstructured formats. We contend that a wealth of knowledge, originally articulated by top-tier human experts at the doctoral level, remains ``locked'' within these complex documents. By deploying a robust parsing methodology capable of navigating heterogeneous PDF layouts, we effectively unlock and distill this information, subjecting the extracted content to a multi-tier quality filtering and deduplication process. This effort culminated in a premier dataset of more than 1.5 millon figure--caption--context triplets derived from over 251k articles, capturing more than 5 million sub-figures with precise spatial localization. OmniScience contains 4.3 billion tokens, including 1.9B image tokens and 2.4B text tokens. It spans over 10 major scientific disciplines, ranging from biology and materials science to ecology and mathematics. The dataset exhibits unprecedented visual heterogeneity and serves as a foundational resource for AI for Science.

To further bridge the semantic gap between visual representations and scientific text, we introduce a dynamic model-routing re-captioning pipeline. This pipeline strategically orchestrates a suite of frontier MLLMs, including Gemini-3-Pro, GPT-5, and Qwen3-VL-235B~\cite{qwen3vl}, to generate dense, self-contained descriptions. It synthesizes visual features with original captions and all corresponding in-text paragraphs, enriching the multi-level information content of the metadata. This process effectively infuses the recaptions with both the specialized knowledge of human experts and the advanced reasoning capabilities of state-of-the-art models. To ensure high data fidelity, we implement a rigorous quality control framework. This framework includes cross-modal deduplication, multi-tier quality filtering, and hallucination detection, complemented by an LLM-based judging mechanism aligned with human expert standards. The resulting textual descriptions are both semantically rich and factually precise, significantly enhancing cross-modal alignment for downstream tasks. Specifically, this re-captioning approach improves the image–text similarity score on Qwen3-VL-Reranker-8B~\cite{qwen3vlembedding} from $0.769$ to $0.956$, while increasing the average caption length from $106.2$ to $360.6$ words.

Furthermore, to demonstrate the effectiveness of OmniScience for MLLM training, we finetune Qwen2.5-VL-3B~\cite{qwen25vl} on OmniScience and evaluate the quality of generated captions on the test set using an LLM-as-a-judge protocol, showing substantial gains over both the pre-trained model and baseline finetuning settings. We further propose a novel caption-based QA evaluation paradigm that treats generated captions as reliable visual proxies for assessing visual understanding. Empirical results validate the effectiveness of this protocol, with models finetuned on OmniScience achieving consistent performance improvements across established benchmarks, including absolute gains of $+0.378$ on MM-MT-Bench~\cite{agrawal2024pixtral}, $+0.140$ on MMMU~\cite{mmmu}, and $+0.083$ on MSEarth~\cite{zhao2025msearth}.

We summarize the primary contributions of this work as follows:

\begin{itemize}[leftmargin=1.5em]
    \item We introduce OmniScience, a large-scale, high-fidelity scientific multi-modal dataset comprising 1.5 million figure–caption–context triplets from 25 top-tier open-access journals and preprints across 10 major scientific disciplines.
    \item We present a framework for enriching scientific image captions from figure–caption–context triplets. Our approach generates dense, self-contained descriptions that improve image comprehension, as shown by strong benchmark performance and a novel caption QA evaluation.
\end{itemize}

%% file: sec/2_related_work.tex
\section{Related Work}
\label{sec:relatedwork}

\subsection{Scientific Figure Understanding}
The development of benchmarks for scientific figure understanding has evolved from synthetic visualizations to complex, multi-modal figures from real-world literature. 
Early efforts established foundational visual reasoning tasks using synthetically generated charts in datasets like FigureQA~\cite{figureqa}, DVQA~\cite{dvqa}, and PlotQA~\cite{plotqa}. To address the lack of real-world complexity, subsequent research shifted to authentic charts extracted from the arXiv repository, with benchmarks such as SciCap~\cite{hsu2021scicap,yang2024scicap+}, ChartQA~\cite{chartqa}, CharXiv~\cite{wang2024charxiv}, and SciFiBench~\cite{roberts2024scifibench} introducing tasks like captioning and complex question-answering. Recognizing that science relies on more than just charts, the focus then broadened to encompass a wider array of visual modalities. MMSCI~\cite{li2024mmsci} provided a large-scale, multidisciplinary benchmark of high-quality figures from the structured HTML of Nature Communications. Similarly, datasets like MMArxiv~\cite{arxivqa} and SciMMIR~\cite{wu2024scimmir} were curated to include heterogeneous imagery such as schematics, microscopy images, and experimental results, often linking them to corresponding article text to enable deeper, multi-modal reasoning. While these datasets represent significant progress, our work extends this frontier by processing the highly heterogeneous PDF landscape to achieve broader coverage than is possible from structured sources. Furthermore, we introduce a novel recaptioning process that leverages large multimodal models to systematically enhance the fidelity of the extracted figure-caption pairs, resulting in a dataset of superior quality.

\begin{table*}[t]
\caption{Comparison with previous scientific figure datasets. \textbf{OmniScience} covers a broader range of scientific disciplines, a wider set of high-quality open-access sources, and higher-quality standards. It provides richer captions and multi-level data, enabling deeper visual-textual grounding and stronger source authority. $^*$Diverse figures include charts, diagrams, microscopy images, molecular structures, and reaction equations. \textsuperscript{\dag}OmniScience contains 1.5M main figure–caption–context triplets with 5M+ localized sub-figures.}
\centering
\small
\begin{tabular}{lcccccccc}
\toprule
\textbf{Dataset} & \textbf{Source Type} & \textbf{Subject} & \textbf{\# Figures} & \textbf{Modality} & \textbf{Context} & \textbf{Subfig. Info} & \textbf{Recaption} \\
\midrule
FigureQA~\cite{figureqa} & Synthetic & N/A & 140K & Charts & \textcolor{red}{\ding{55}} & \textcolor{red}{\ding{55}} & \textcolor{red}{\ding{55}} \\
DvQA~\cite{dvqa} & Synthetic & N/A & 300K & Charts & \textcolor{red}{\ding{55}} & \textcolor{red}{\ding{55}} & \textcolor{red}{\ding{55}} \\
SciCap~\cite{yang2024scicap+} & arXiv & CS & 414K & Charts & \textcolor{green}{\ding{51}} & \textcolor{red}{\ding{55}} & \textcolor{red}{\ding{55}} \\
SciFiBench~\cite{roberts2024scifibench} & arXiv & CS & 6K & Charts & \textcolor{red}{\ding{55}} & \textcolor{red}{\ding{55}} & \textcolor{red}{\ding{55}} \\
M-Paper~\cite{mpaper} & arXiv & CS & 350K & Diverse & \textcolor{green}{\ding{51}} & \textcolor{red}{\ding{55}} & \textcolor{red}{\ding{55}} \\
CharXiv~\cite{wang2024charxiv} & arXiv & Multi. & 2.3K & Charts & \textcolor{red}{\ding{55}} & \textcolor{red}{\ding{55}} & \textcolor{red}{\ding{55}} \\
MMArxiv~\cite{arxivqa} & arXiv & Multi. & 6.4M & Diverse & \textcolor{green}{\ding{51}} & \textcolor{red}{\ding{55}} & \textcolor{red}{\ding{55}} \\
MMSCI~\cite{li2024mmsci} & Nat. Commun. & Natural Sci. & 742K & Diverse & \textcolor{green}{\ding{51}} & \textcolor{red}{\ding{55}} & \textcolor{red}{\ding{55}} \\
\midrule
\rowcolor{blue!5} \textbf{OmniScience} & \textbf{25 High-Impact OA} & \textbf{10 Disciplines} & \textbf{1.5M / 5M$^\dag$} & \textbf{Diverse$^*$} & \textcolor{green}{\ding{51}} & \textcolor{blue}{\textbf{BBox + Type}} & \textcolor{blue}{\textbf{MLLM-Enriched}} \\
\bottomrule
\end{tabular}
\label{tab:datasets_comparison}
\end{table*}

\subsection{Scientific Document Mining}

The curation of large-scale scientific vision-language datasets hinges on the effective extraction of information from scholarly publications. However, traditional repositories like PubMed Central are often subject to domain-specific constraints and frequently suffer from significant noise, including low-resolution imagery, misaligned figure--text pairs, and the prevalence of non-informative graphical elements~\cite{hsu2021scicap,li2024mmsci}. While parsing structured HTML or LaTeX files offers cleaner metadata~\cite{wang2024charxiv,arxivqa}, these formats are restricted to specific journals or preprints, leaving the vast majority of scientific knowledge locked within heterogeneous PDF layouts. Despite the advancement of general-purpose document AI tools such as MinerU~\cite{wang2024mineru} and DeepSeek-OCR~\cite{wei2025deepseekocr}, they remain optimized for text-to-markdown conversion and often struggle with the fine-grained spatial alignment of figures and context. This bottleneck is effectively addressed by Uni-Parser~\cite{fang2025uniparser}, which employs a group-based layout design to ensure the robust binding of images with their respective captions and in-text references. By achieving superior accuracy across diverse scientific disciplines, Uni-Parser serves as a vital complement to HTML-based curation, finally rendering the assembly of encyclopedic, high-fidelity scientific datasets feasible.

\subsection{Quality Assessment for Image Caption}

Evaluating the fidelity of large-scale vision-language data remains a persistent challenge. Traditional n-gram metrics, such as BLEU~\cite{papineni2002bleu} and CIDEr~\cite{vedantam2015cider}, primarily assess surface-level fluency and often fail to capture the deep semantic nuances inherent in scientific imagery. While embedding-based measures like CLIPScore~\cite{hessel2021clipscore} have long served as the standard for web-scale filtering~\cite{laion}, recent studies indicate they frequently overlook domain-specific details, such as precise data points or intricate structural relationships in scientific figures~\cite{whenandwhy}. To overcome these limitations, the field has recently shifted toward leveraging the advanced reasoning capabilities of MLLMs. Next-generation multimodal embeddings~\cite{gunther2025jina} and rerankers~\cite{wang2025jina}, such as the latest Qwen3-VL-Ranker~\cite{qwen3vlembedding}, have demonstrated superior performance in cross-modal alignment by effectively distilling knowledge from state-of-the-art models. Concurrently, an emerging paradigm in both industry and academia seeks to evaluate data by converting images into descriptive captions that serve as functional surrogates for downstream tasks~\cite{yang2025captionqa, fang2025uniparser}. Furthermore, reference-free assessment frameworks like G-Eval~\cite{G-eval}, which utilize LLM-as-a-judge methodologies, offer a more nuanced understanding of alignment and factual integrity. These evolving techniques provide a robust foundation for our multi-tier evaluation framework.

%% file: sec/3_data_curation.tex
\section{Dataset Curation}
\label{sec:data_curation}

\begin{figure*}[ht]
\centering
\begin{subfigure}[h]{0.4\textwidth}
    \centering
    \includegraphics[width=\textwidth]{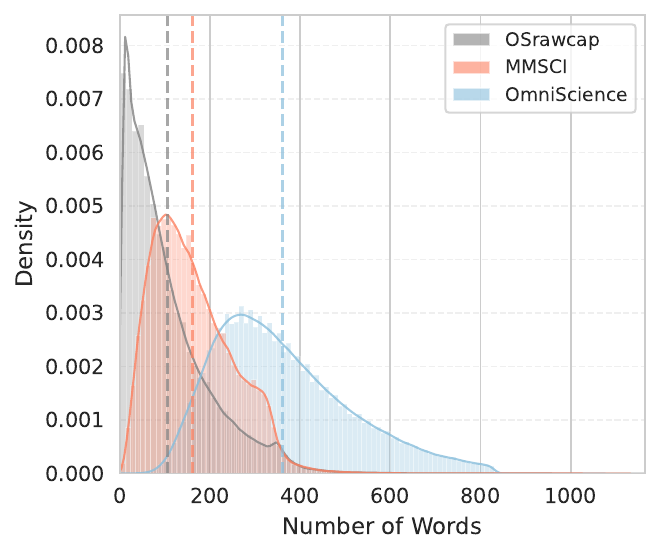}
    \caption{Comparison of Caption Length Distributions}
    \label{fig:sub1}
\end{subfigure}
\hspace{0.5cm}
\begin{subfigure}[h]{0.4\textwidth}
    \centering
    \includegraphics[width=\textwidth]{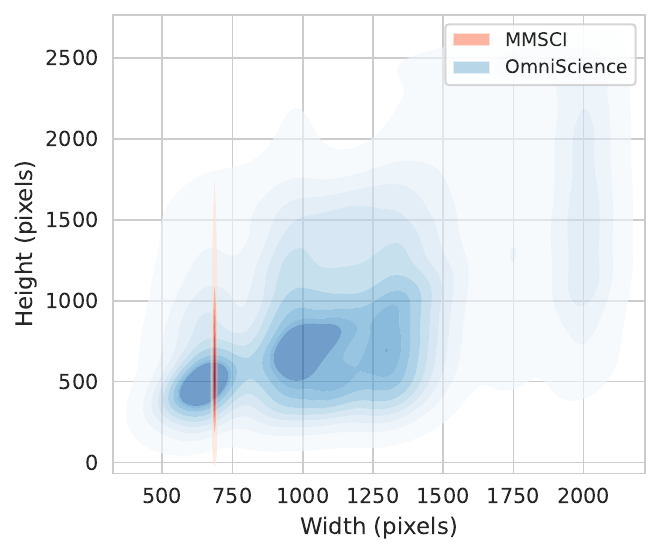}
    \caption{Comparison of Image Resolution Distributions}
    \label{fig:sub2}
\end{subfigure}
\caption{Caption length and image resolution comparison between the OmniScience and MMSCI datasets. (a) Probability density distributions of caption lengths (in words) for both datasets.
(b) Two-dimensional density distribution of image resolutions (width and height in pixels). The pronounced vertical line in the MMSCI distribution is a processing artifact, as all images in its public release were constrained to a fixed width of 685 pixels.}
\label{fig:distribution_comparison}
\end{figure*}

\subsection{Source Data Collection}

\noindent
\textbf{High-Quality Open-Access Journals.}\quad To ensure the dataset's authority and academic rigor, our primary source consists of high-quality, peer-reviewed journal articles. We exclusively focused on Open Access (OA) repositories to align with open-science principles. The collection, finalized as of September 25, 2025, covers a broad spectrum of scientific domains—ranging from life sciences and chemistry to physics and materials science—to ensure a diverse disciplinary distribution. The strategy included:
\begin{itemize}[leftmargin=1.5em]
    \item \textbf{\textit{Nature Communications}:} We incorporated the complete MMSCI corpus~\cite{li2024mmsci} and performed targeted crawling to ensure full coverage of this premier multidisciplinary venue.
    \item \textbf{Global Disciplinary Coverage:} To maximize breadth, we integrated top-tier generalist journals and the top 20--30 ranked journals (by impact factor) across each major scientific discipline, maintaining a baseline of vetted, high-fidelity scientific content.
\end{itemize}

\noindent
\textbf{Strategic Preprint Integration.}\quad To supplement the source distribution and capture emerging research trends or unique visual styles not yet present in traditional journals, we collected preprints from servers such as arXiv and bioRxiv (cutoff May 30, 2025). To maintain the high quality of the dataset, we applied a stringent dual-gate filter: we only retained papers that possessed at least one citation and were simultaneously within the top 20\% by download volume. This was followed by a secondary semantic assessment using GPT-4o~\cite{gpt4o} based on titles and abstracts. Furthermore, to capture the most influential recent developments in AI, we included a manually curated set of computer-science preprints from the Hugging Face Trending Papers\footnote{https://huggingface.co/papers/trending}.

\noindent
\textbf{Metadata Acquisition.}\quad 
For each document, we retrieved core metadata (DOI, open-access status, download links) via Unpaywall\footnote{https://unpaywall.org/} and respected each publication's licensing terms, excluding items with prohibitive licenses.
We augmented these records with disciplinary information from OpenAlex\footnote{https://openalex.org/}
 via DOI lookups. Because OpenAlex labels proved at times inconsistent, overly granular, or missing, we implemented a pipeline that uses GPT-4o to map each article (title plus original subject data) into a standardized discipline taxonomy. This produced a uniform, high-quality classification across the dataset.

\noindent
\subsection{Figure-caption-context Extraction}
After collecting the source high quality PDF files, we implemented a multi-stage pipeline to systematically extract structured information from each document. This process included deduplication, figure-caption pair extraction, context retrieval, and sub-figure identification.

\noindent
\textbf{PDF-level Deduplication.}\quad We eliminated redundant documents by cross-referencing Digital Object Identifiers (DOIs) and normalized titles. The normalization process involved converting text to lowercase, removing punctuation, and collapsing multiple whitespace sequences into single spaces. In cases of metadata collisions (e.g., identical DOIs or near-duplicate titles), we established a clear priority hierarchy: peer-reviewed journal versions were favored over preprints, and the most recent iterations were prioritized over earlier versions.

\noindent
\textbf{Figure-Caption Pair Extraction.}\quad 
To address the challenges of extracting structured data from heterogeneous PDF layouts, our pipeline leverages the \textit{Uni-Parser} framework for robust scientific document parsing. Specifically, we utilize the \textit{Uni-Parser-LD} module to perform initial layout detection and precisely localize figure-caption groups. Building upon these detections, we perform {sub-figure localization and legend extraction, segmenting primary figures into their constituent panels. We integrate PaddleOCR~\cite{cui2025paddleocr30technicalreport} to identify rasterized identifiers (e.g., 'a', 'b') within image regions. These extracted legends, along with sub-captions parsed from the main text, are then mapped to their respective panels through a set of proprietary heuristic matching rules. Crucially, to resolve structural discontinuities such as cross-column and cross-page separations, we implemented a heuristic merging strategy that operates on the logical reading order provided by Uni-Parser. By analyzing the serialized sequence of document elements, our system can reconstruct the association between figures and their corresponding captions even when they are physically decoupled by column or page boundaries—a scenario occurring in approximately 3\% of our corpus. Finally, the extracted text undergoes rigorous sanitization to remove control symbols while carefully preserving essential scientific notations and chemical symbols. To ensure maximum data integrity, we adopt a precision-oriented filtering strategy by excluding problematic cases, such as figure-caption matching failures, one-to-many mapping conflicts, and instances with low-confidence detection or recognition scores. Evaluation on a 500-PDF benchmark demonstrates that our extraction pipeline achieves 100\% precision in figure-caption pair extraction; a detailed breakdown of our validation protocol and performance metrics is provided in Appendix~\ref{appendix:extraction}.

\noindent
\textbf{Contextual Content Extraction.}\quad To associate each figure with its textual context, we developed an automated matching procedure. First, we applied regular expression rules to extract key identifiers (e.g., ``Figure 1'', ``Fig. 2a'', , ``Schema 1A'') from the caption text. The same rules were then used to locate all mentions of these identifiers within the main body of the article. When a reference was matched, the entire natural paragraph containing it was extracted and stored as the figure's associated context.

\subsection{Data Filtering}
During the extraction of raw structured data, we performed an initial round of filtering to ensure the accuracy of the extracted triplets. Building upon this step, we further implemented a comprehensive cleaning and filtering pipeline to improve the overall quality and consistency of the final dataset. This stage consists of two parallel processes that target the visual and textual components, respectively.

\noindent
\textbf{Figure Filtering.}\quad The extracted images underwent a two-stage refinement process.
First, to address the prevalence of near-duplicate figures within individual articles, we conducted a secondary deduplication using image p-Hash with 0.965 similarity threshold, removing a small fraction of redundant visuals. 
Second, we applied quality-based filtering to exclude low-resolution images (smaller than $200 \times 200$ pixels) and those with extreme aspect ratios that typically indicate formatting errors. 

\noindent
\textbf{Caption Filtering.}\quad The corresponding captions were subjected to a series of quality-control filters.
First, we removed figure–caption pairs in which the caption was excessively short and lacked supplementary context; specifically, instances with ten or fewer words and no associated in-text context were discarded. 
Second, we applied a sentence completeness check to filter out fragmented captions. Captions were required to begin with a capital letter and end with standard terminal punctuation (e.g., a period or semicolon) to be considered syntactically complete. 
Finally, we sanitized the caption text using regular expressions to remove extraneous metadata, such as embedded DOI links and copyright notices, which are not part of the descriptive content.

\subsection{Dynamic Model Routing for Recaptioning}
The raw extracted captions, despite being written by human experts, vary widely in quality and completeness. Owing to heterogeneous journal sources and diverse authoring styles, many captions include loosely related or irrelevant text, while a large fraction are overly concise and abstract (e.g., “reaction scheme of compound 1 and compound 2”), offering little concrete description of the visual content.
To standardize the format and substantially enhance the descriptive richness of the dataset, we develop a recaptioning pipeline that orchestrates a suite of state-of-the-art MLLMs to produce dense, self-contained descriptions. Each caption is generated by jointly conditioning on the original figure, its human-written caption, and the associated in-text context, leading to more informative and semantically aligned supervision for multimodal model training.

Inspired by recent work on efficient model orchestration such as Octopus v4~\cite{chen2024octopus}, we designed our pipeline not around a single, monolithic model, but around an efficient and performance-aware routing system. The central goal was to dynamically allocate the most suitable LLM for a given figure type, thereby maximizing generation quality while optimizing computational expense. Our router operates as a cascaded decision-making system, directing each figure-caption pair to a specific model or a pool of models based on its characteristics.

The model-routing strategy is driven by a fine-grained categorization of figures, combined with cues extracted from both visual content and the original human-written captions. We first classify each figure by subject area and visual type, and then dynamically assign it to the most suitable model according to the complexity of visual understanding required, the expected recaptioning quality, and practical constraints such as API cost and service load balancing.

Figures with complex experimental measurements or fine-grained visual patterns (e.g., SEM, NMR, XPS), chemical structures, or many subfigures are preferentially routed to the Gemini family (e.g., Gemini-3-Pro and Gemini-2.5-Pro~\cite{gemini}), which performs particularly well on data-intensive scientific imagery; samples with long contextual paragraphs are handled by Gemini-3-Flash to support extended context at lower cost. Charts, plots, and schematics emphasizing quantitative or logical relations are routed to GPT-5 models, while simpler figures are handled by cost-efficient models (Qwen3-VL-235B~\cite{qwen3vl}, Gemini-2.5-Flash, GPT-5 mini) to scale within budget. To ensure robustness, we use a cascading fallback: requests fall back to Gemini-3-Flash-Preview and, if needed, a local Qwen3-VL-235B upon API failure or refusal, mitigating interruptions from instability or moderation. Model routing statistics are reported in Appendix~\ref{appendix:statistics}.

Recaptioning is performed by jointly conditioning on the original image, the original caption, and all in-text paragraphs that reference the figure, producing dense and self-contained descriptions. The prompt templates used in this process are detailed in Appendix~\ref{appendix:prompts}.

\subsection{Quality Filtering for Recaption}
\label{sec:quality_filter}

To ensure factual accuracy and mitigate hallucinations or repetition, we implemented a three-stage quality assurance pipeline:

\noindent
\textbf{Heuristic and Repetition Filtering.}\quad 
We automatically flag captions exceeding 830 words to prevent infinite generation loops. A hybrid approach—combining rule-based analysis with LLM-based self-consistency checks—is used to identify and remove structural and semantic redundancies.

\noindent
\textbf{Visual-Text Consistency via Multimodal Fact-Checking.}\quad 
A VLM-based "Fact-Checker" performs triangular verification by comparing the candidate description against the source image, the original author caption, and the article context. This stage specifically targets three hallucination types: pattern extensions (logical but non-existent sequences), visual fabrications (textual mentions not present in the image), and unsourced specifics (unverifiable data points).

\noindent
\textbf{Automated Regeneration Loop.}\quad 
Any sample flagged as low-quality enters a regeneration loop. The system retries the generation by feeding specific failure reasons (e.g., "hallucination detected") back into the prompt to guide the model toward a valid output.

\subsection{Dataset comparison}
OmniScience substantially extends prior scientific figure datasets in both scale and data fidelity. As shown in Tab.~\ref{tab:datasets_comparison} and Fig.~\ref{fig:distribution_sources}, whereas existing benchmarks are typically restricted to specific domains or single venues, OmniScience aggregates figures from 25 high-impact journals and preprint archives, offering broader disciplinary coverage and more diverse visual styles. Unlike most prior datasets, OmniScience further provides fine-grained sub-figure bounding boxes and associated in-text context paragraphs, enabling deeper multimodal reasoning. Beyond annotation richness, OmniScience establishes a new standard for caption quality and visual fidelity: the caption length distribution (Fig.~\ref{fig:distribution_comparison}(a)) indicates that our recaptioning pipeline yields substantially more descriptive captions (mean length: 360.55 words) than both the original extracted captions and those in prior datasets such as MMSCI, while the resolution distribution (Fig.~\ref{fig:distribution_comparison}(b)) shows that OmniScience preserves native high-resolution figures without the uniform downscaling and aspect-ratio distortion observed in MMSCI (fixed width of 685 pixels). OmniScience provides a more faithful and high-quality representation of multi-modal scientific data.

%% file: sec/4_experiment.tex
\section{Experiments}
\label{sec:exps}

\subsection{Effectiveness of Recaptioning}
\label{sec:exp_recaption}

To quantify the effect of recaptioning on image--caption semantic alignment, we conduct an intrinsic cross-modal retrieval evaluation on the held-out OmniScience test set. We compare raw figure--caption pairs with our recaptioned figure--caption pairs generated by our pipeline, and measure how well captions align with their corresponding images in a shared embedding space.

We employ Qwen3-VL-Reranker-8B~\cite{qwen3vlembedding} to compute fine-grained relevance scores between images and captions. Compared to embedding based dual-tower similarity models, the reranker adopts a single-tower cross-encoder that jointly encodes image–caption pairs and explicitly models cross-modal interactions using cross attention, yielding more precise instance-level semantic grounding. Embedding models are typically used for efficient coarse-grained retrieval, whereas reranker models provide more accurate pairwise relevance estimation for multi-modal similarity. Therefore, we adopt the higher-precision reranker to compute image–caption similarity scores in our evaluation. Compared to CLIP Score~\cite{hessel2021clipscore}, which only supports short-text inputs and is much smaller in scale, these metrics are of far greater practical relevance for scientific multimodal evaluation.

\begin{figure}[ht]
    \centering
    \begin{subfigure}[b]{0.9\linewidth}
        \includegraphics[width=\linewidth]{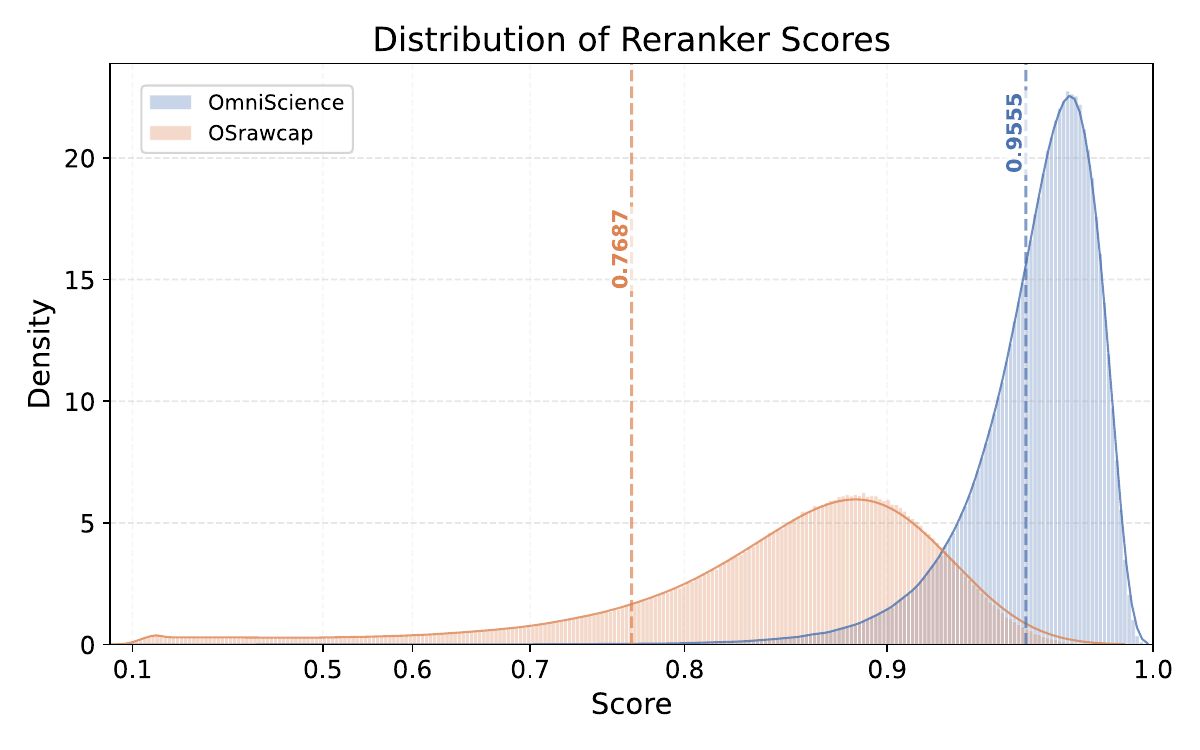}
    \end{subfigure}
    \caption{Distribution of multi-modal reranker scores for captions from OSrawcap (OmniScience raw captions) and OmniScience (re-captioned). The horizontal axis uses a non-linear scale.}
    \label{fig:caption_reranker_1}
\end{figure}

We provide quantitative comparisons in Figure~\ref{fig:caption_reranker_1} to demonstrate the improved quality of OmniScience captions over the original raw captions. The figure shows the distribution of cross-modal relevance scores computed by Qwen3-VL-Reranker-8B for image–caption pairs constructed from raw captions and OmniScience re-captions, respectively.

As shown in Fig.~\ref{fig:caption_reranker_1}, OmniScience captions exhibit a substantially right-shifted and more concentrated score distribution with a significantly higher mean relevance score, indicating stronger and more consistent semantic grounding between images and captions. In contrast, raw captions produce lower and more variable scores, reflecting their frequent brevity, abstraction, or incompleteness. This quantitative evidence supports that OmniScience provides higher-quality, self-contained captions for multi-modal learning.

\subsection{Performance Benchmarking Across Caption Data}
\label{sec:exp_mllm_finetune}

To further investigate whether the enhanced semantic alignment provided by OmniScience captions improves multi-modal representation learning, we conduct a series of fine-tuning experiments. Specifically, we evaluate whether training a Multi-modal Large Language Model (MLLM) on OmniScience captions yields superior cross-modal alignment compared to using the original raw captions. To this end, we construct a validation set by randomly sampling 500 images from each of the 10 major disciplines, totaling 5,000 images. We perform full-parameter fine-tuning of the Qwen2.5-VL-3B model~\cite{qwen25vl} on various image captioning datasets. Unless otherwise specified, the detailed training configurations follow the recipe provided in Appendix~\ref{appendix:recipe} and are maintained consistently across all subsequent experiments.

\begin{figure}[h]
    \centering
    \begin{subfigure}[b]{0.9\linewidth}
        \includegraphics[width=\linewidth]{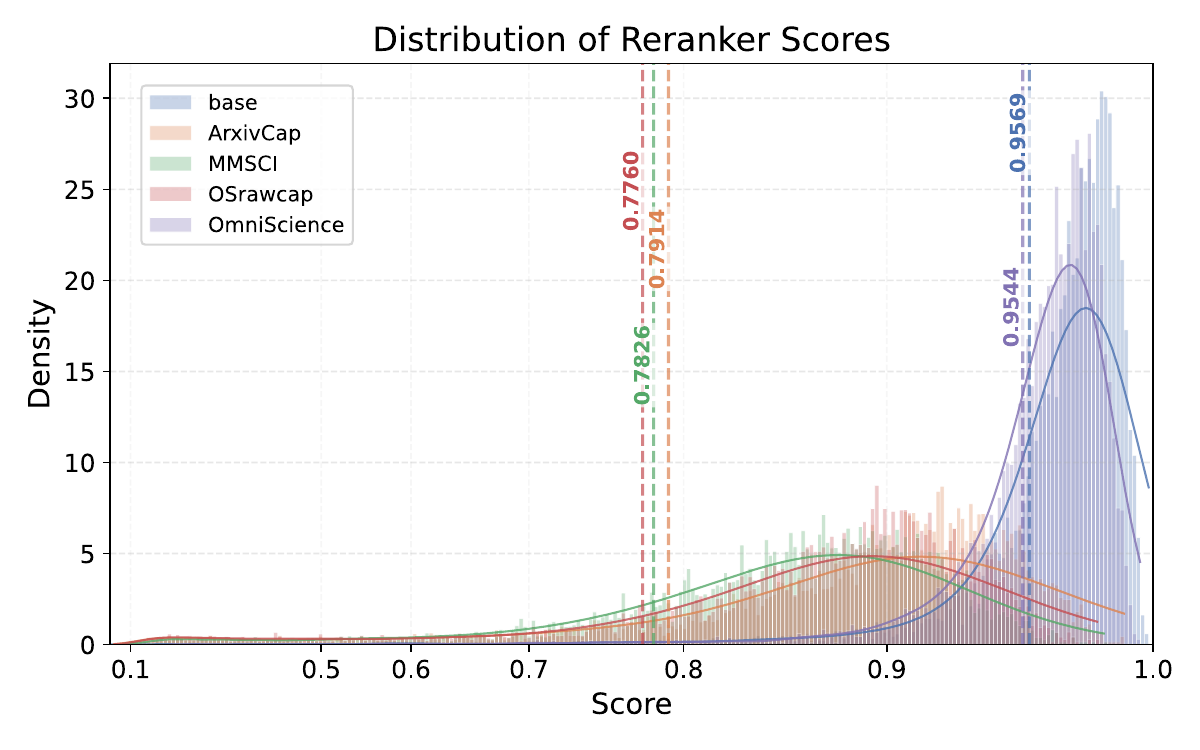}
    \end{subfigure}
    \caption{Distribution of multi-modal reranker scores for captions generated by Qwen2.5-VL-3B fine-tuned on different datasets, evaluated on the OmniScience validation set.}
    \label{fig:caption_reranker_for_models}
\end{figure}

We present the comparative results of caption generation quality in Figure~\ref{fig:caption_reranker_for_models}. We observe that models fine-tuned on external datasets (e.g., MMSCI, ArxivCap) or raw captions can display suboptimal performance, degrading to levels substantially below the baseline. In contrast, the model fine-tuned on the OmniScience achieves significantly higher multi-modal reranker scores, validating the effectiveness of our recaptioning methodology. Notably, the unfine-tuned base model also attains high reranker scores, likely due to the strong baseline capability of Qwen2.5-VL and potential data homology between the Qwen-based generator and the Qwen-based reranker used for evaluation.

\subsection{Quality Assessment via LLM-as-a-Judge}
\label{sec:exp_llm_judge}

After training the model, to enable a fine-grained evaluation of caption quality beyond coarse similarity, we implement an LLM-as-a-Judge framework using a mean ensemble of state-of-the-art MLLMs, including Qwen3-VL-235B-A22B-Thinking~\cite{qwen3vl} and Seed-1.5VL (doubao-1-5-thinking-vision-pro-250428)~\cite{seed1_5vl}. To ensure reliability, we define four rigorous evaluation dimensions and average the scores from both judges.

\begin{itemize}[leftmargin=1.5em]
    \item \textbf{Language Fluency:} Assesses the text's coherence, grammar, and natural flow.
    \item \textbf{Information Consistency:} Evaluates how consistently the caption text matches the observable content in the image.
    \item \textbf{Key Information Accuracy:} Measures whether the caption accurately captures the main subjects, relationships, and important details visible in the image.
    \item \textbf{Detail Level:} Compares the generated caption to the original reference to assess the comprehensiveness of the description.
\end{itemize}

Each dimension is scored on a 1–5 scale. On 300 expert-annotated samples, LLM-as-a-Judge achieves a Quadratic Weighted Kappa of 0.831 with human ratings, indicating strong agreement and high reliability.

\begin{figure}[ht]
    \centering
    \begin{subfigure}[b]{0.9\linewidth}
        \includegraphics[width=\linewidth]{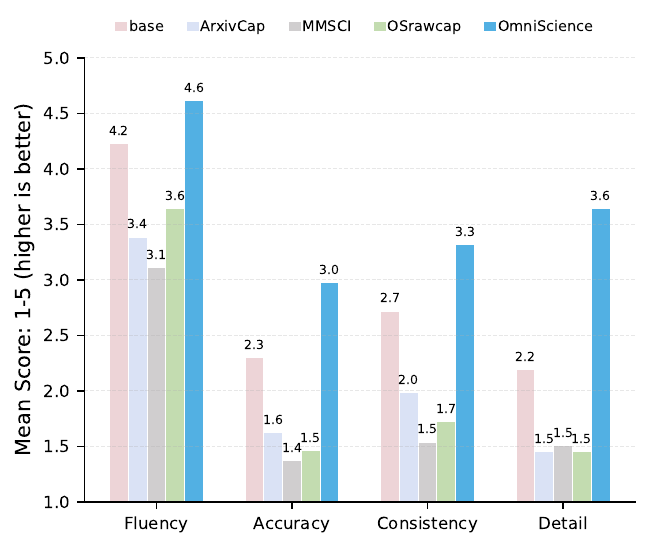}
    \end{subfigure}
    \caption{Quality assessment of generated captions, evaluated by LLM-as-a-judge. Higher indicates better quality.}
    \label{fig:assessment_of_LLM}
\end{figure}

As illustrated in Figure~\ref{fig:assessment_of_LLM}, the model fine-tuned on OmniScience consistently outperforms all baseline models across every dimension. Notably, significant improvements are observed in \textit{Information Consistency}, \textit{Key Information Accuracy} and \textit{Detail Level}. This suggests that by conditioning on global article context, our pipeline successfully mitigates the "referential ambiguity" (e.g., phrases like "as shown in the left panel" appearing without context) common in raw captions, resulting in self-contained and factually grounded descriptions.

\subsection{Caption QA as a Proxy Task}
\label{sec:exp_capqa}

While intrinsic assessment measures caption quality, we further evaluate their functional utility using a "Caption QA" bottleneck framework. This experiment tests the hypothesis that a high-quality scientific caption can serve as a sufficient information proxy for the image, enabling a model to reason about scientific content without direct visual input. This methodology supports the argument that image captioning, far beyond a rudimentary descriptive task, constitutes a critical nexus for multimodal alignment whose role in bridging visual perception and logical reasoning has been significantly undervalued \cite{yang2025captionqa}.

\noindent\textbf{Experimental Setup:} Following the same protocol as Section~\ref{sec:exp_recaption}, we train separate Qwen2.5-VL-3B~\cite{qwen25vl} models on different scientific image caption datasets. We then transform established VQA benchmarks, AI2D~\cite{ai2d} (diagrams), MMMU~\cite{mmmu} (expert-level), MM-MT-Bench~\cite{agrawal2024pixtral} (multi-turn), and MSEarth~\cite{zhao2025msearth} (remote sensing), into caption tasks. In each task, the visual placeholder (\texttt{<image>}) is replaced by the caption generated by the corresponding model. A unified reasoning engine (GPT-4o-mini) answers questions using only these captions.

\noindent\textbf{Quantitative Results:} Table~\ref{tab:capqa_res} shows that models trained on raw captions (e.g., Qwen-MMSCI, OmniSci-rawcap) often underperform, sometimes even below the baseline, suggesting that human-written captions in scientific papers can be too context-dependent or concise to function standalone. In contrast, the model trained on OmniScience captions (after recaption pipeline) consistently achieves superior performance, with substantial absolute gains: +0.585 on MM-MT-Bench, +0.297 on MMMU, and +0.129 on MSEarth, relative to the raw-caption counterpart.

These results demonstrate that our recaptioning pipeline goes beyond simple text rephrasing, effectively distilling complex visual information into dense, textually accessible captions, thereby substantially improving the performance of downstream scientific MLLM training.

\begin{table}[ht]
\centering
\caption{Caption QA performance of Qwen2.5-VL-3B models fine-tuned on different caption datasets, evaluated on VQA using GPT-4o-mini. Higher scores indicate that training with this dataset enhances scientific image reasoning capabilities.}
\label{tab:capqa_res}
\footnotesize
\begin{tabular}{lcccc}
\toprule
\textbf{Training Set} & \textbf{AI2D} & \textbf{MM-MT-Bench} & \textbf{MMMU} & \textbf{MSEarth} \\
\midrule
None      & \textbf{0.612} & 0.338 & 0.230 & 0.096 \\
MMSCI~\cite{li2024mmsci}   & 0.544 & 0.181 & 0.133 & 0.067 \\
MMArxiv~\cite{arxivqa}     & 0.527 & 0.093 & 0.110 & 0.057 \\
\midrule
OmniSci-rawcap             & 0.555 & 0.131 & 0.073 & 0.050 \\
\rowcolor{blue!5}\textbf{OmniScience}  & 0.603 & \textbf{0.716} & \textbf{0.370} & \textbf{0.179} \\
\bottomrule
\end{tabular}
\end{table}

%% file: sec/5_conclusion.tex
\section{Conclusion}
We introduce OmniScience, a large-scale multi-modal dataset comprising 1.5 million figure-caption-context triplets across 10 major scientific disciplines. Central to our contribution is a dynamic model-routing recaptioning pipeline that synthesizes visual features with human-authored context to generate dense, self-contained descriptions. This systematic enhancement of data fidelity effectively "unlocks" expert-level knowledge, boosting image-text embedding similarity from 0.769 to 0.956 and tripling the average descriptive length. By addressing the critical semantic gap in scientific image understanding, OmniScience enables substantial performance leaps on expert-level benchmarks. We expect this high-fidelity resource to serve as a cornerstone for future AI scientists, bridging the divide between complex visual evidence and large-scale multi-modal comprehension.

%% file: sec/appendix.tex
\clearpage
\appendix
\onecolumn

\section{Quality Assessment of Figure--Caption--Context Extraction}
\label{appendix:extraction}

In this appendix, we present a quantitative evaluation and statistical analysis of the extraction quality of figure--caption--context triplets in OmniScience. Our evaluation prioritizes precision over exhaustive recall, reflecting the design principle that data quality is more critical than raw quantity for training multimodal foundation models.

We uniformly sample 500 PDF documents across all data sources to construct the evaluation set and conduct manual verification for statistical analysis. After the first-stage online quality control, the recall of figure--caption pairs reaches 67.56\%, while the accuracy is 100.00\%. For context paragraphs, we achieve a recall of 99.91\% with an accuracy of 100.00\%. For structurally challenging cases, the recall of cross-column figure--caption associations is 55.56\%, and the recall of cross-page associations is 63.05\%. On average, each figure is associated with 2.56 context paragraphs in OmniScience, providing rich complementary information beyond the original caption.

Despite partial recall loss in complex layouts, the consistently perfect accuracy demonstrates that the proposed pipeline is highly reliable and effectively prevents mismatched figure--text associations from being introduced into the dataset. These results validate that OmniScience offers high-fidelity figure--text--context alignment suitable for training precision-sensitive multimodal models.

\section{Additional Statistics and Visual Analysis}
\label{appendix:statistics}

\begin{figure}[ht]
\centering
\includegraphics[width=0.9\columnwidth]{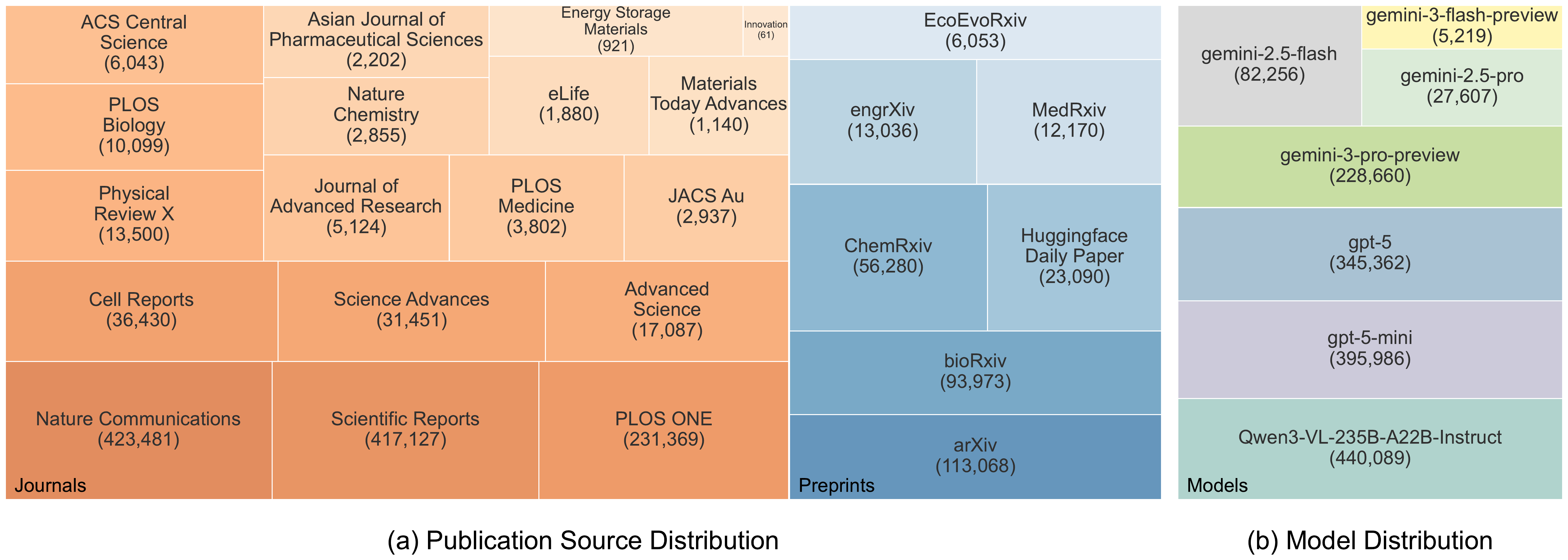}
\caption{Source distribution overview. (a) Distribution of figure–caption pairs across journals and preprints. (b) Distribution of models employed in the re-captioning pipeline.}
\label{fig:distribution_sources}
\end{figure}

\begin{figure}[h]
    \centering
    \begin{subfigure}[b]{0.41\textwidth}
        \centering
        \includegraphics[width=\textwidth]{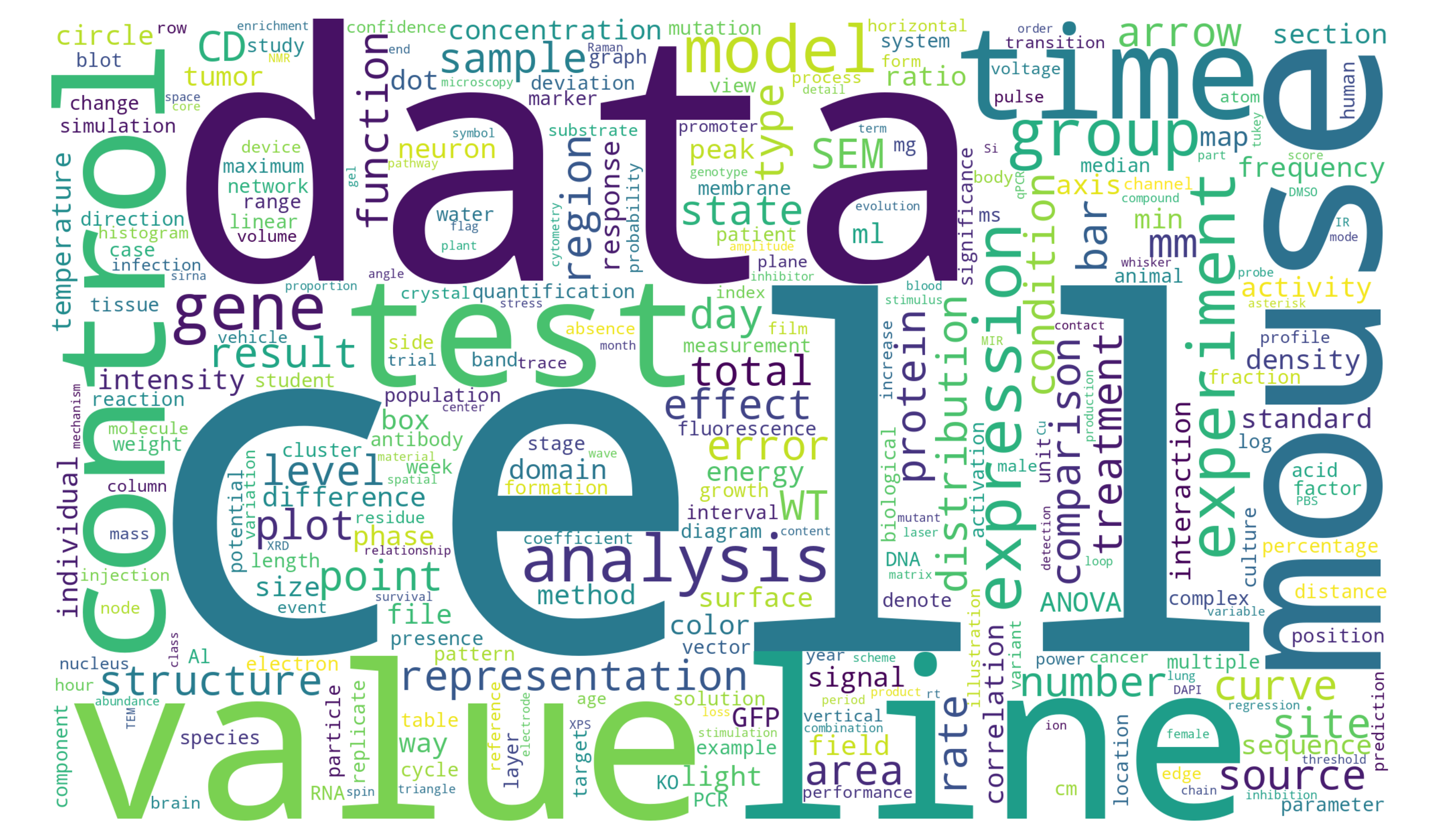}
        \caption{Original Captions}
        \label{fig:wordcloud_raw}
    \end{subfigure}
    \hspace{0.3cm}
    \begin{subfigure}[b]{0.41\textwidth}
        \centering
        \includegraphics[width=\textwidth]{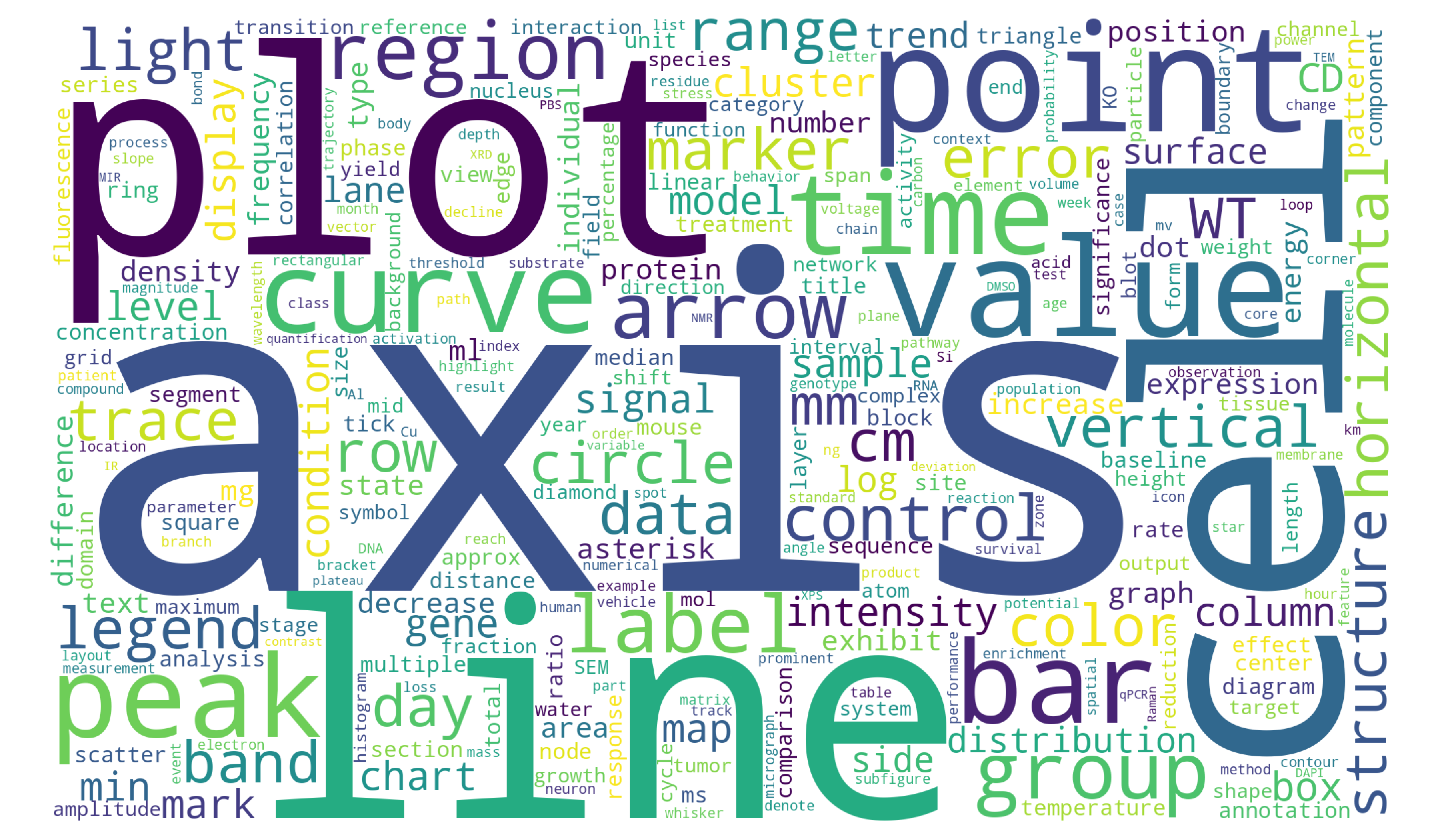}
        \caption{Re-captioned Descriptions}
        \label{fig:wordcloud_recap}
    \end{subfigure}
    
    \caption{Visualizing lexical distributions via word clouds for (a) original captions and (b) re-captioned descriptions. After filtering generic nouns, the re-captioned data exhibits a higher density of fine-grained, content-bearing terminology.}
    \label{fig:wordcloud_comparison}
\end{figure}

We provide a visualization of the source distribution, and list the corresponding data sources in Fig.~\ref{fig:distribution_sources}. We also conduct a statistical analysis of lexical distributions for both the original figure captions and the re-captioned descriptions. To facilitate a meaningful comparison, we extract and analyze word frequencies at the noun level and visualize the resulting distributions using word clouds. During preprocessing, non-informative or generic nouns (e.g., demonstrative placeholders such as this) are systematically removed to reduce semantic noise and highlight content-bearing terminology. As shown in Fig.~\ref{fig:wordcloud_comparison}, after re-captioning, the descriptions focus more strongly on fine-grained details.

\section{Training and Inference Recipe for Image Captioning Models}
\label{appendix:recipe}

For all image captioning models, we perform full-parameter fine-tuning using the \texttt{LLaMA-Factory} framework. The models are trained on a cluster of 8$\times$ NVIDIA A100 GPUs with FlashAttention-2. We employ the AdamW optimizer with a cosine learning rate scheduler. Detailed training hyperparameters and inference configurations are summarized in Table~\ref{tab:hyperparams}.

\begin{table}[ht]
\centering
\caption{Hyperparameters for model training and inference.}
\label{tab:hyperparams}
\begin{tabular}{lc}
\hline
\textbf{Configuration} & \textbf{Value} \\ \hline
\textit{Training Settings} & \\
Fine-tuning Method & full-parameter \\
Precision & bfloat16 \\
Learning Rate & $5 \times 10^{-6}$ \\
Batch Size (per device) & 2 \\
Gradient Accumulation Steps & 2 \\
Number of Epoch & 1 \\
Learning Rate Scheduler & cosine \\
Warmup Ratio & 0.03 \\
Weight Decay & 0.1 \\
Max Gradient Norm & 1.0 \\
Adam $\beta_1$ & 0.9 \\
Adam $\beta_2$ & 0.95 \\ \hline
\textit{Inference Settings} & \\
Temperature & 0.7 \\
Top-$p$ & 0.9 \\ \hline
\end{tabular}
\end{table}


\section{Prompts}
\label{appendix:prompts}

\subsection{Re-caption Prompt}

\begin{tcolorbox}[
title=Re-caption Prompt, 
colback=white, 
colframe=black!70, 
arc=5pt,
fonttitle=\bfseries\ttfamily,
breakable,
enhanced,
pad at break=0pt,
bottomrule at break=0pt,
toprule at break=0pt
]

\textbf{Role}

You are an expert scientific researcher and data annotator. Your task is to generate a refined, scientifically dense and detailed description (``re-caption'') for the provided image, leveraging the Raw Caption and Context.

\textbf{Objective}

Create a JSON output containing a single string that captures the \textbf{scientifically significant} information from the image. Focus on visual evidence that supports scientific conclusions (trends, anomalies, key measurements, structural topology) while ignoring low-level, non-semantic clutter.

\textbf{Input Data}

\begin{itemize}[label=--, leftmargin=2em]
    \item \textbf{[Image]}: <image>
    \item \textbf{[RAW\_CAPTION]}: \{raw\_caption\_string\}
    \item \textbf{[CONTEXT]}: \{context\_list\_of\_string\}
\end{itemize}

\textbf{Detailed Guidelines}

\begin{enumerate}[label=\arabic*., leftmargin=2em]
    \item \textbf{Information Selection Strategy (The ``Salience'' Filter)}:
    \begin{itemize}[label=--, leftmargin=2em]
        \item \textbf{Context-Guided Identification:} Use the [RAW\_CAPTION] and [CONTEXT] to identify \textit{what} the object is (e.g., ``Figure 3 shows the FTIR spectrum of Sample A'').
        \item \textbf{Visual-Based Verification:} Describe \textit{how} it looks based strictly on visual evidence. (You may use scientific information from the raw caption and context, but \textbf{strictly on visual evidence})
        \item \textbf{Key Information vs. Noise:}
        \begin{itemize}[label=--, leftmargin=2em]
            \item \textbf{DO EXTRACT:} Global trends (increasing, decreasing, saturation), critical inflection points (maxima, minima, plateaus), specific values of semantic importance (peaks, outliers, scale bar values), and relationships between variables.
            \item \textbf{DO NOT EXTRACT:} Every single tick mark on an axis, standard grid lines, minor background noise, or purely decorative font styles unless they obscure the data.
        \end{itemize}
    \end{itemize}

    \item \textbf{Category-Specific Instructions}:
    \begin{itemize}[label=--, leftmargin=2em]
        \item \textbf{Charts/Plots:} Write the precise \textbf{underlying data tables} if possible. For complex charts or situations where it is difficult to convert to data tables, describe the relationship between axes, state the overall trend (e.g., ``exponential decay,'' ``linear correlation''), extract numerical values for key features (e.g., ``peak absorbance at 550 nm,'' ``saturation efficiency of $\sim$20\%''), mention error bars if significant.
        \item \textbf{Spectra (XRD, NMR, FTIR, etc.):} Focus on defining features: specific peak positions (wavenumbers/2-theta), relative intensities (strong/weak), and peak shapes (broad/sharp). (Charts and Spectra are important data for scientific analysis which need more detailed caption for downstream data analysis.)
        \item \textbf{Microscopy (SEM/TEM/AFM):} Extract the scale (e.g., ``scale bar: 500 nm''). Describe morphology, size distribution, texture, and surface features.
        \item \textbf{Schematics/Flowcharts:} Describe the topological flow or reaction mechanism. Focus on the logical path (A $\rightarrow$ B $\rightarrow$ C) and structural changes rather than colors or box shapes.
        \item For multi-subplot figures, adopt a ``general--specific--general'' structure. If any subplot conveys sufficiently important information, it must be described in detail.
    \end{itemize}

    \item \textbf{Style \& Tone}:
    \begin{itemize}[label=--, leftmargin=2em]
        \item \textbf{Directness:} Start immediately with the subject or the finding. \textbf{Strictly NO} ``This image shows,'' ``The recaption of image is,'' ``Image Description'' ``The figure illustrates,'' ``We can see'' or ``Figure/Schema x shows xxx''. (DO NOT INCLUDE FIGURE ID, because it can not be obtained from image)
        \item \textbf{Complete detailing but concise language:} Use dense Academic English. Merge visual descriptions with their scientific implications derived from context.
    \end{itemize}

    \item \textbf{Output Format}:
    \begin{itemize}[label=--, leftmargin=2em]
        \item Only return the refined image description, no any other texts.
    \end{itemize}
\end{enumerate}

\textbf{Examples of ``Salient'' vs. ``Too Detailed / Too brief (info loss)''}

\textbf{Bad (Too Detailed/Mechanical):}

``The X-axis has ticks at 0, 10, 20, 30, 40, 50. The Y-axis has ticks at 0.0, 0.5, 1.0. There is a blue line that starts at 0,0 and goes to 10, 0.2, then to 20, 0.4\ldots''

\textbf{Bad (Too brief / Lack of specifics scientific numerical information or conclusion):}

``\ldots (c) Barchart for resnet validation results in ImageNet. Red for model A and Blue for model B''. \ldots

\textbf{Good (Scientifically Salient):}

``Linear dependence of conductivity on temperature, ranging from 0 to 50$^\circ$C. Conductivity increases steadily from negligible values at 0$^\circ$C to a peak of $\sim$1.0 S/cm at 50$^\circ$C, indicating semiconductor behavior. \ldots''

\textbf{Output Format}

Return strictly on this format:

\begin{verbatim}
Your refined image description string here, used for describe image in detail 
for scientific analysis. For dense numerical information, you can use markdown 
format such as tables or lists.
\end{verbatim}

\end{tcolorbox}


\subsection{Quality Assessment Prompt}

\begin{tcolorbox}[
title=Quality Assessment Prompt, 
colback=white, 
colframe=black!70, 
arc=5pt,
fonttitle=\bfseries\ttfamily,
breakable,
enhanced,
pad at break=0pt,
bottomrule at break=0pt,
toprule at break=0pt
]

\textbf{Role}

You are a rigorous Multimodal Fact-Checker. Your task is to verify a generated text description against three sources of ground truth:
\begin{enumerate}[label=\arabic*., leftmargin=2em]
    \item \textbf{The Image} (Visual evidence)  <image>
    \item \textbf{The Original Caption} (Basic ground truth) \{raw\_caption\_string\}
    \item \textbf{The Context/Article} (Textual evidence) \{context\_list\_of\_string\}
\end{enumerate}

\textbf{Objective}

Identify \textbf{Hallucinations} in the ``Candidate Description''.

A hallucination is defined as specific information (entities, numbers, text sequences, quantitative data) that \textbf{cannot be found in ANY of the three provided sources}, but is fabricated by the model---often due to pattern completion or over-inference.

\textbf{The ``Overkill'' Prevention Mechanism (Nuance Rules)}

To avoid false positives (overkill), adhere to these distinctions:

\begin{enumerate}[label=\arabic*., leftmargin=2em]
    \item \textbf{ALLOWED (Not Hallucination):}
    \begin{itemize}[label=--, leftmargin=2em]
        \item Synthesis: Combining info from the Image and the Article. (e.g., Image shows ``Level 1'', Article mentions ``Level 4 exists''. Output ``Levels 1 and 4'' is VALID).
        \item Generalization: Describing the scene using broader terms (e.g., calling ``Level 1'' a ``menu item'').
        \item Implicit Knowledge: Common sense deductions (e.g., acknowledging it is a ``digital interface'' even if not explicitly written).
    \end{itemize}

    \item \textbf{FORBIDDEN (Hallucination):}
    \begin{itemize}[label=--, leftmargin=2em]
        \item \textbf{Pattern Extension (Primary Target):} The source shows/mentions ``A, B, C'', but the output adds ``D, E'' purely because it follows the sequence, WITHOUT any evidence in the Image or Article.
        \item \textbf{Visual Fabrication:} Claiming the \textit{image shows} something that is only mentioned in the text (or nowhere at all).
        \item \textbf{Unsourced Specifics:} Specific dates, names, or numbers that appear nowhere in the sources.
    \end{itemize}
\end{enumerate}

\textbf{Critical Rules}

\begin{enumerate}[label=\arabic*., leftmargin=2em]
    \item \textbf{Multi-Source Verification:} Information is valid if it appears in ANY of the three sources (Image $\cup$ Caption $\cup$ Context).
    \item \textbf{Visual Claims Require Visual Evidence:} If the candidate claims something is ``visible in the image'', it MUST be visually present.
    \item \textbf{Pattern Extension Detection:} Focus on sequences (numbers, letters, steps, levels) that continue beyond source evidence.
    \item \textbf{Context Integration:} Use Article text to validate claims that aren't visually present but are mentioned in context.
\end{enumerate}

\textbf{Examples}

\textbf{Example 1 (Pattern Extension Hallucination)}

\textbf{Sources:}
\begin{itemize}[label=--, leftmargin=2em]
    \item Image: Shows ``Stage 1, Stage 2''
    \item Article: ``This process is efficient.'' (No mention of Stage 3)
\end{itemize}

\textbf{Candidate:} ``The image displays the workflow including Stage 1, Stage 2, and continues to Stage 3.''

\textbf{Analysis:} Stage 3 is not visible in the image and not mentioned in the article. The model hallucinated the sequence continuation.

\textbf{Example 2 (Valid Cross-Modal Synthesis)}

\textbf{Sources:}
\begin{itemize}[label=--, leftmargin=2em]
    \item Image: Shows ``Level 1''
    \item Article: ``Players can unlock Level 2 and Level 3 later in the game.''
\end{itemize}

\textbf{Candidate:} ``The game currently shows Level 1, but Level 2 and Level 3 are available features.''

\textbf{Analysis:} Although Level 2 and 3 are not in the image, they are supported by the Article text. This is a valid synthesis, not a hallucination.

\textbf{Example 3 (Visual Misattribution)}

\textbf{Sources:}
\begin{itemize}[label=--, leftmargin=2em]
    \item Image: Shows a simple chart
    \item Article: ``The report discusses quarterly growth from Q1 to Q4''
\end{itemize}

\textbf{Candidate:} ``The image shows quarterly growth data from Q1 to Q4 with trend lines.''

\textbf{Analysis:} If trend lines aren't visible in the image, this is visual fabrication even if Q1--Q4 are mentioned in the article.

\textbf{Output Format}

Return ONLY a JSON object:

\begin{verbatim}
{
  "has_hallucination": true/false,
  "hallucination_type": "Pattern_Extension / Data_Fabrication / 
                         Visual_Misattribution / None",
  "severity_score": 1-5,
  "reason": "Brief explanation of the hallucination or 
             'No hallucination detected'"
}
\end{verbatim}

\end{tcolorbox}

\clearpage

\subsection{LLM-as-a-judge Prompts}

\begin{tcolorbox}[
title=Language Fluency Evaluation Prompt, 
colback=white, 
colframe=black!70, 
arc=5pt,
fonttitle=\bfseries\ttfamily,
breakable,
enhanced,
pad at break=0pt,
bottomrule at break=0pt,
toprule at break=0pt
]

\textbf{Role}

You are an expert evaluator for text quality. Analyze the generated caption text to assess its language fluency.

\textbf{Generated Caption Text to Assess:}

\{model\_prediction\}

\textbf{Evaluation Criteria}

\textbf{Language Fluency} (Score 1--5): How fluent and logically coherent is the language in the caption text?

\begin{itemize}[label=--, leftmargin=2em]
    \item 5 = Perfectly fluent, natural language flow, excellent grammar and structure
    \item 4 = Very fluent with minor issues that don't affect readability
    \item 3 = Moderately fluent, some awkward phrasing but understandable
    \item 2 = Poor fluency, multiple grammar/structure issues affecting readability
    \item 1 = Very poor fluency, difficult to understand
\end{itemize}

\textbf{Output Format}

Please provide your assessment in the following JSON format:

\begin{verbatim}
{
    "language_fluency": {
        "score": <integer 1-5>,
        "reasoning": "<brief explanation>"
    }
}
\end{verbatim}

\end{tcolorbox}

\begin{tcolorbox}[
title=Information Consistency Evaluation Prompt, 
colback=white, 
colframe=black!70, 
arc=5pt,
fonttitle=\bfseries\ttfamily,
breakable,
enhanced,
pad at break=0pt,
bottomrule at break=0pt,
toprule at break=0pt
]

\textbf{Role}

You are an expert evaluator for image captioning quality. Analyze the provided image and the generated caption text to assess information consistency.

<image>

\textbf{Generated Caption Text to Assess:}

\{model\_prediction\}

\textbf{Evaluation Criteria}

\textbf{Information Consistency} (Score 1--5): How consistent is the caption text with what can be observed in the image?

\begin{itemize}[label=--, leftmargin=2em]
    \item 5 = Perfectly consistent, all described information matches the image
    \item 4 = Highly consistent with only minor discrepancies
    \item 3 = Moderately consistent, some information doesn't match the image
    \item 2 = Poorly consistent, significant mismatches between caption and image
    \item 1 = Very inconsistent, most information contradicts or doesn't appear in image
\end{itemize}

\textbf{Output Format}

Please provide your assessment in the following JSON format:

\begin{verbatim}
{
    "information_consistency": {
        "score": <integer 1-5>,
        "reasoning": "<brief explanation>"
    }
}
\end{verbatim}

\end{tcolorbox}

\clearpage

\begin{tcolorbox}[
title=Key Information Accuracy Evaluation Prompt, 
colback=white, 
colframe=black!70, 
arc=5pt,
fonttitle=\bfseries\ttfamily,
breakable,
enhanced,
pad at break=0pt,
bottomrule at break=0pt,
toprule at break=0pt
]

\textbf{Role}

You are an expert evaluator for image captioning quality. Analyze the provided image and the generated caption text to assess key information accuracy.

<image>

\textbf{Generated Caption Text to Assess:}

\{model\_prediction\}

\textbf{Evaluation Criteria}

\textbf{Key Information Accuracy} (Score 1--5): How accurately does the caption capture the key/important information visible in the image?

\begin{itemize}[label=--, leftmargin=2em]
    \item 5 = All key information is accurately captured (main subject, important details, relationships)
    \item 4 = Most key information is accurate with minor omissions or imprecisions
    \item 3 = Some key information is accurate, but important elements are missing or imprecise
    \item 2 = Key information is mostly inaccurate or missing
    \item 1 = Key information is completely wrong or absent
\end{itemize}

\textbf{Output Format}

Please provide your assessment in the following JSON format:

\begin{verbatim}
{
    "key_information_accuracy": {
        "score": <integer 1-5>,
        "reasoning": "<brief explanation>"
    }
}
\end{verbatim}

\end{tcolorbox}

\begin{tcolorbox}[
title=Detail Level Evaluation Prompt, 
colback=white, 
colframe=black!70, 
arc=5pt,
fonttitle=\bfseries\ttfamily,
breakable,
enhanced,
pad at break=0pt,
bottomrule at break=0pt,
toprule at break=0pt
]

\textbf{Role}

You are an expert evaluator for image captioning quality. Compare the generated caption with the original captions to assess the level of detail.

\textbf{Original Captions (reference):}

\{original\_caption\}

\textbf{Generated Caption Text to Assess:}

\{model\_prediction\}

\textbf{Evaluation Criteria}

\textbf{Detail Level} (Score 1--5): How detailed and comprehensive is the generated caption compared to the original captions?

\begin{itemize}[label=--, leftmargin=2em]
    \item 5 = Comprehensive detail covering all important aspects mentioned in original captions
    \item 4 = Good level of detail, covers most important aspects
    \item 3 = Moderate detail, covers some aspects but misses others
    \item 2 = Sparse detail, only basic information provided
    \item 1 = Very little detail, lacks substantive description
\end{itemize}

\textbf{Output Format}

Please provide your assessment in the following JSON format:

\begin{verbatim}
{
    "detail_level": {
        "score": <integer 1-5>,
        "reasoning": "<brief explanation>"
    }
}
\end{verbatim}

\end{tcolorbox}


\newpage
\section{OmniScience Showcases of Re-captioning}
\label{appendix:case}

\subsection{OmniScience Case Study 1}
\begin{figure}[!htpb]
  \centering
  \includegraphics[width=\linewidth]{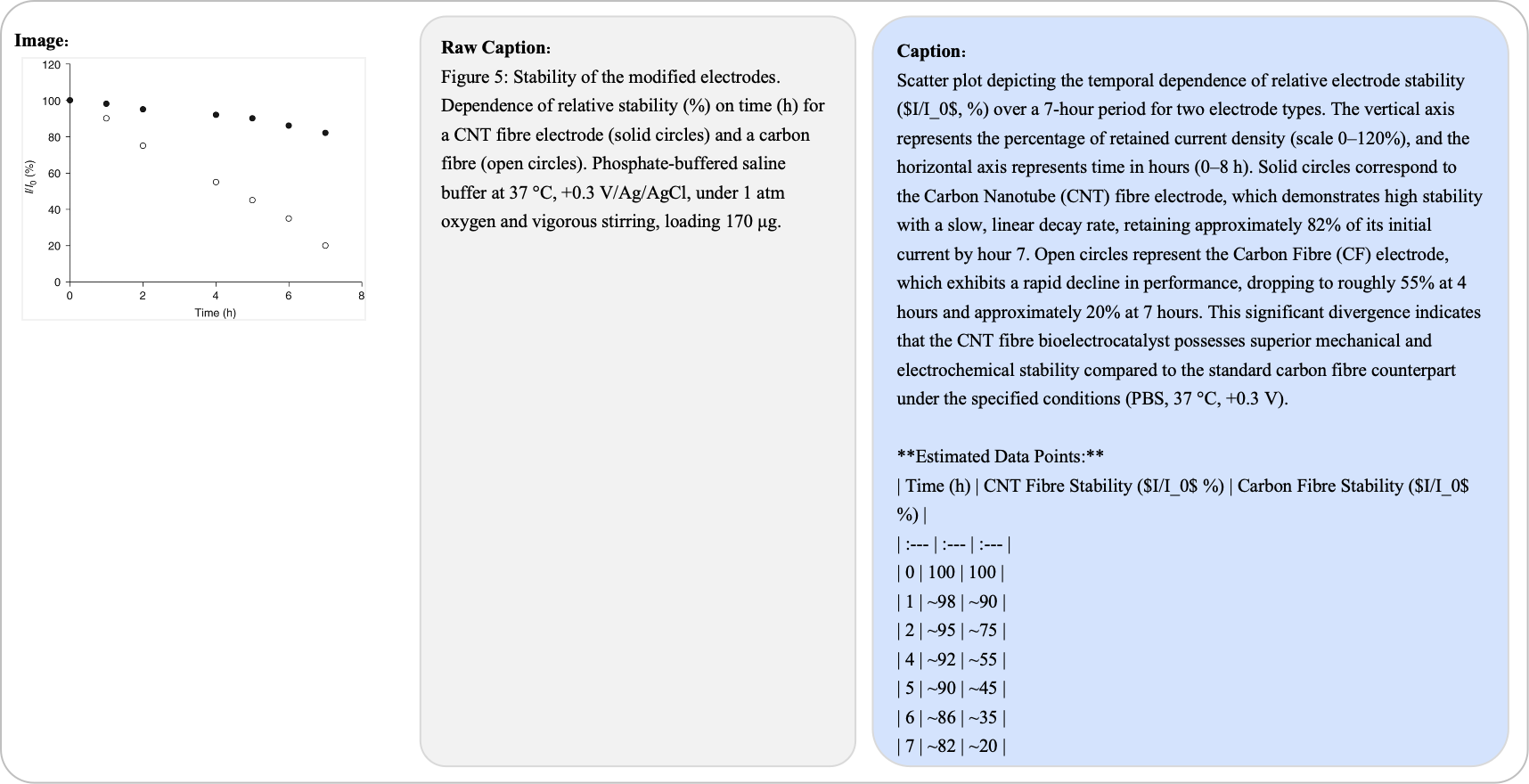}
\end{figure}

\subsection{OmniScience Case Study 2}
\begin{figure}[!htpb]
  \centering
  \includegraphics[width=\linewidth]{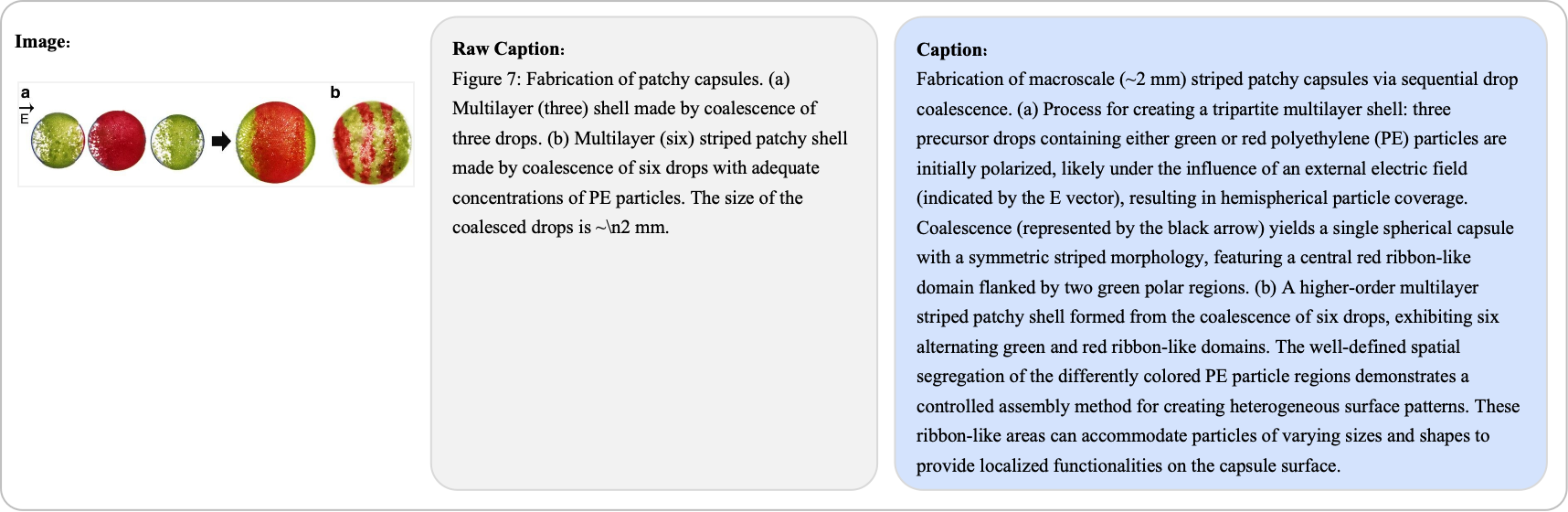}
\end{figure}

\newpage

\subsection{OmniScience Case Study 3}
\begin{figure}[!htpb]
  \centering
  \includegraphics[width=\linewidth]{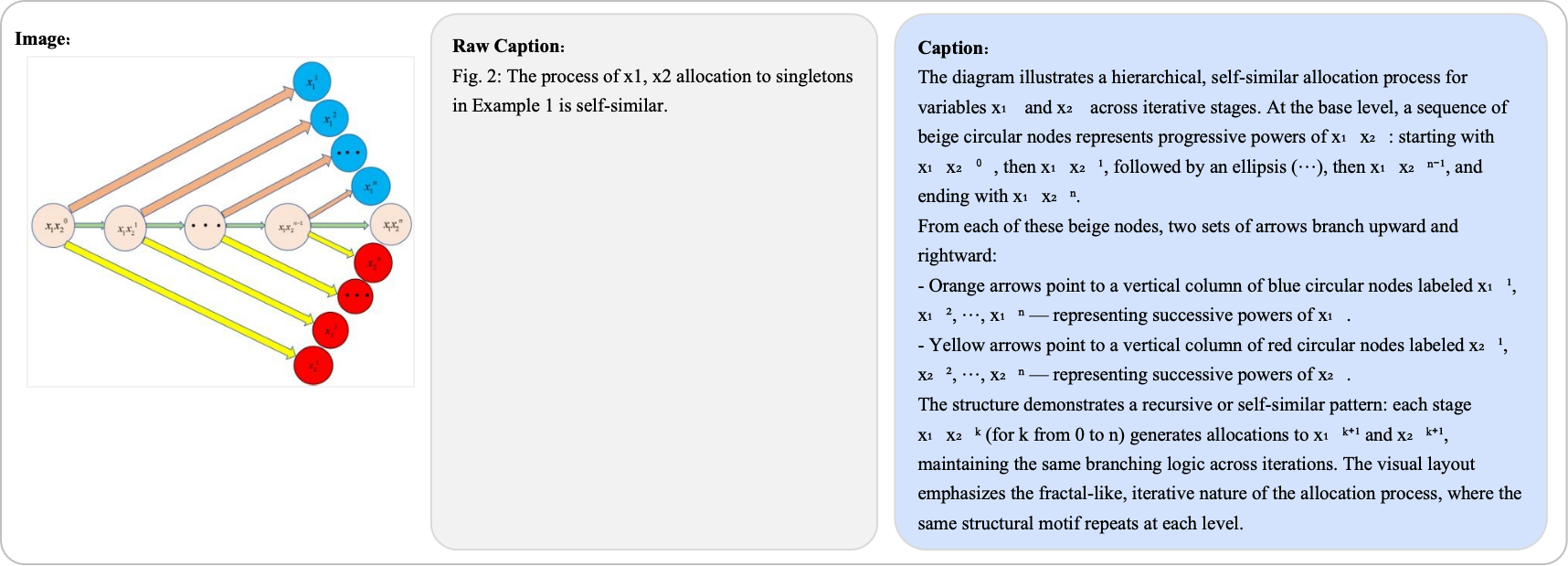}
\end{figure}

\subsection{OmniScience Case Study 4}
\begin{figure}[!htpb]
  \centering
  \includegraphics[width=\linewidth]{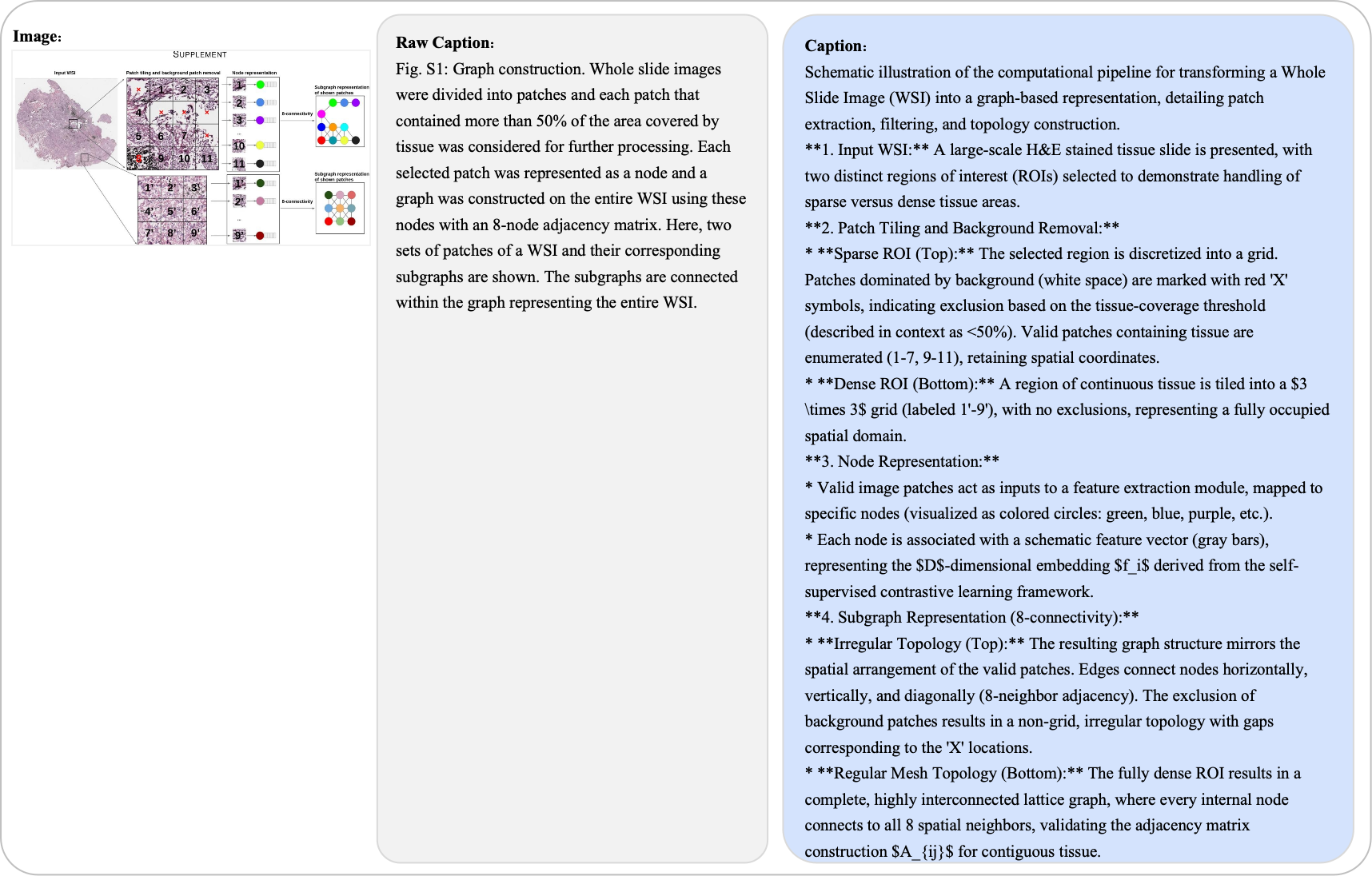}
\end{figure}

\newpage

\subsection{OmniScience Case Study 5}
\begin{figure}[!htpb]
  \centering
  \includegraphics[width=\linewidth]{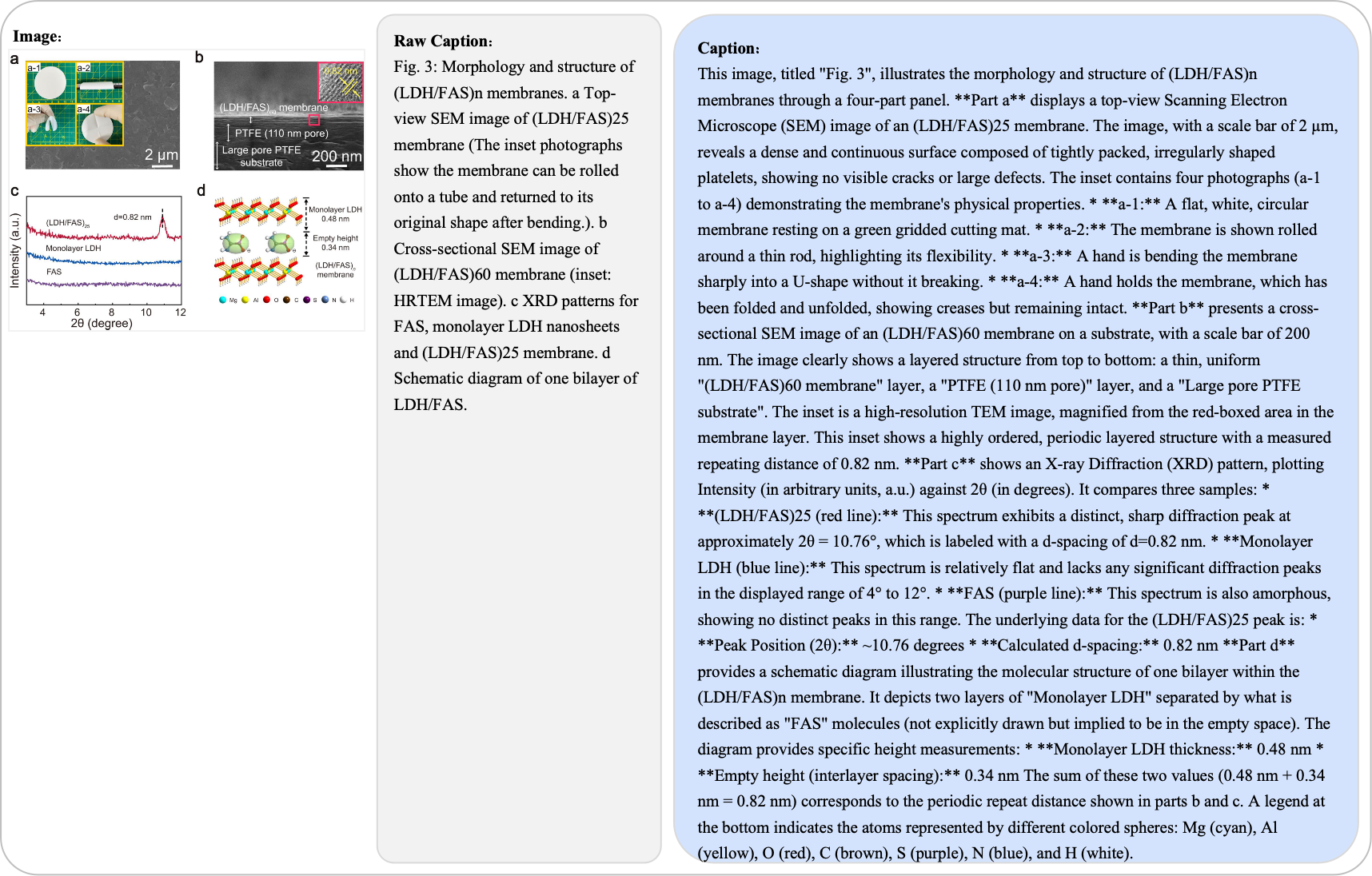}
\end{figure}

\subsection{OmniScience Case Study 6}
\begin{figure}[!htpb]
  \centering
  \includegraphics[width=\linewidth]{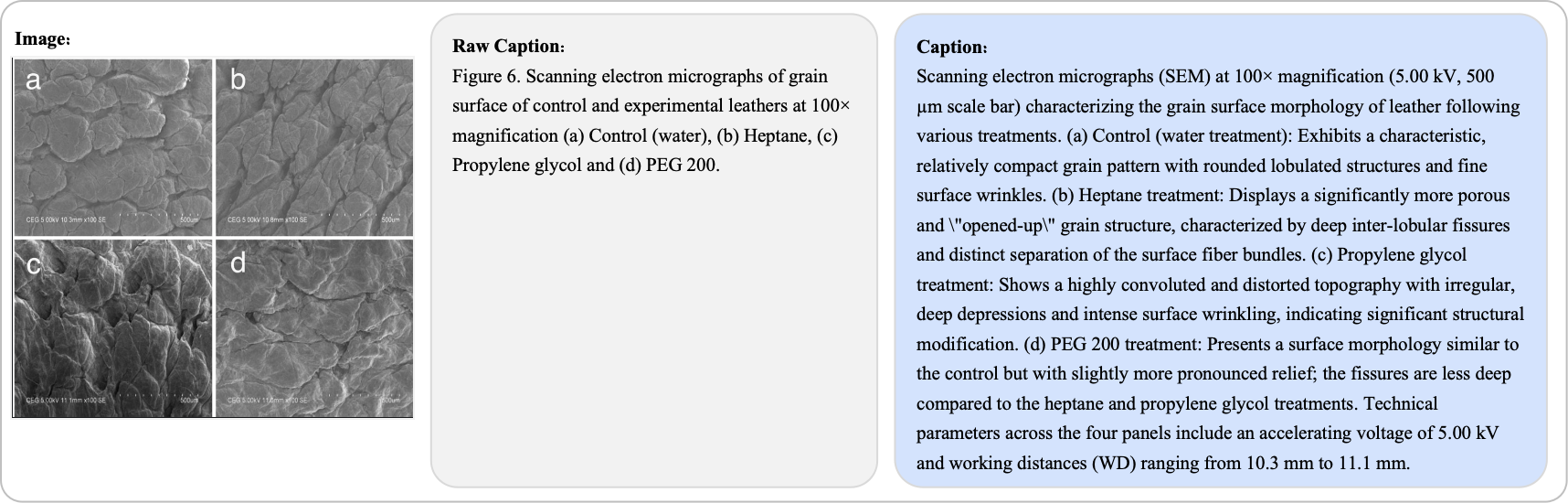}
\end{figure}